\documentclass[10pt,twocolumn]{article}
\usepackage{simpleConference}

\usepackage{graphicx}
\usepackage[tight]{subfigure}
\usepackage{float}

\begin{document}

\title{QRMODA and BRMODA: Novel Models for Face Recognition Accuracy in Computer Vision Systems with Adapted Video Streams}
\author{Hayder R. Hamandi and Nabil J. Sarhan \\
Wayne State University\\
Detroit, MI}

\maketitle

\begin{abstract}
A major challenge facing Computer Vision systems is providing
the ability to accurately detect threats and recognize subjects and/or
objects under dynamically changing network conditions. 
We propose two novel models that characterize 
the face recognition accuracy in terms of video encoding parameters.
Specifically, we model the accuracy in terms of video resolution, quantization, and actual bit rate. 
We validate the models using two distinct video datasets and a large image dataset by conducting $1,668$ experiments that involve simultaneously varying combinations of encoding parameters. 
We show that both models hold true for the {\em deep learning} and statistical based face recognition. 
Furthermore, we show that the models can be used to capture different accuracy metrics, specifically the recall, precision, and F1-score. Ultimately, we provide meaningful insights on the factors affecting the constants of each proposed model.

\end{abstract}

\section{Introduction}
\label{sec:intro}

The video streams in Computer Vision (CV) systems should be adapted dynamically (by changing the video capturing and encoding parameters) to fit the tight resource constraints, including network bandwidth, energy, and storage. 
Therefore, these adaptations lead to various tradeoffs involving the accuracy and the aforementioned constraints. 
The overwhelming majority of studies on CV focused on the development of robust algorithms 
to improve the accuracy in primarily image datasets, 
through statistical \cite{csu} (and references within) and more recently deep learning approaches \cite{googlenet, vgg, mobilefacenets, deepfacedetector} (and references within).

We analyze the behavior of CV accuracy 
focusing primarily on face recognition. 
We make a fundamental contribution by developing two novel models that help in assessing the effect of combining adaptation strategies for the same video stream. 
The first model (QRMODA) characterizes CV accuracy in terms of the spatial resolution and quantization parameter ($Q_p$),  
which we determine as a logistic function of $Q_p$, with the $x_0$ value of the Sigmoid's midpoint being a function of the resolution. 
In contrast, the second model (BRMODA) shows the accuracy in terms of the spatial resolution and actual bitrate. 
We find that the accuracy is equal to the sum of two exponentials of the {\em actual} bitrate,
with the resolution as a multiplicative factor with one exponential.

Furthermore, we validate each model against two different (deep learning and statistical based) approaches of face recognition,  
utilizing two greatly distinct video datasets (Honda/UCSD \cite{Honda} and  DISFA \cite{DISFA1}), and a large image dataset (Labeled Faces in the Wild (LFW) \cite{lfw}). We conduct $1,668$ actual experiments on $99$ videos and $13,233$ images, with $47$ and $5,749$ subjects, respectively.  
Subjects have different gender, ethnicity, and pose variations. 
The results indicate that both proposed models hold true for both face recognition approaches and using different datasets. The results also show that the models can characterize face detection. 
Moreover, the models can be used to characterize different accuracy metrics, specifically recall, precision and F1-score, but we 
focus primarily on recall due to its importance in our particular application. 
We compute the coefficient of determination ($R^2$) to assess the goodness of fit. 
Ultimately, we discuss the factors impacting the constants of each proposed model and how to compute them in actual systems.

The {\bf main contributions} can be summarized as
follows: 
$(1)$ developing two mathematical models of face recognition accuracy
in terms of the main video encoding parameters, 
$(2)$ conducting extensive experiments to analyze the impacts of different combinations of video adaptation techniques on CV accuracy,  
$(3)$ validating the two models using two greatly distinct and diverse video datasets and a large image dataset,
and $(4)$ discussing the factors impacting the constants of each model.

The rest of this paper is organized as follows. Section
\ref{sec:background} provides background information and discusses the related work. Section \ref{sec:performance} shows the development of both proposed models. Subsequently, Section \ref{sec:setup} explains the
experimental setup, and Section \ref{sec:validation} presents and analyzes the 
main results. Section \ref{sec:discussion} provides additional discussion and analysis of the factors impacting the constants of both proposed models.
Finally, conclusions are drawn.

\section{Background Information and Related Work}
\label{sec:background}

\subsection{Face Recognition}
Face recognition is a major CV algorithm in many applications, including  authentication systems, personal photo enhancement, automated video surveillance, and photo search engines.
Face recognition approaches can be classified into two broad categories:
{\em neural-based} and {\em statistical-based}.

Neural-based solutions employ neural networks to classify objects within an input image.
\textit{Convolutional Neural Networks} (CNNs)
are the most widely employed type that has proven strong results in terms of accuracy.
Examples include GoogleNet \cite{googlenet}, VGG \cite{vgg}, and MobileFaceNets \cite{mobilefacenets}.
The performance of VGG, GoogleNet, and SqueezeNet has been benchmarked in terms of verification of accuracy against different types of noise in \cite{grm2017}. 
All these architectures aim at reducing the size of the deep CNN. 
We use FaceNet \cite{facenet} with the architecture model targeted towards a datacenter application. 
FaceNet achieved state-of-the-art performance in face recognition, according to a recent survey \cite{soalfw}.

Face recognition using 
statistical-based algorithms can be classified into two main
categories: 
\textit{appearance-based} and \textit{model-based} \cite{Lu2004}.
The first typically represents images as high-dimensional
vectors and then employs statistical techniques, 
such as {\em Principal Component Analysis} (PCA)
or {\em Linear Discriminant Analysis} (LDA), 
for image vector analysis and feature extraction.  
PCA reduces the size of the $n$-dimensional space
used by the
appearance-based algorithm, ultimately simplifying computational
complexity. 
LDA works similarly but requires more training data.
On the other hand, model-based algorithms require
manual human face model construction to capture facial features, while feature
matching is  achieved using an algorithm, such as {\em Elastic Bunch Graph Matching} (EBGM). 
In real-world situations, only a small number of samples for
each subject are available for training. If a sufficient amount of enough representative data is not
available, Martinez and Kak [30] have shown that the switch from nondiscriminant techniques (e.g.,
PCA) to discriminant approaches (e.g., LDA) is not always warranted and may sometimes lead to
poor system design when small and nonrepresentative training data sets are used.
For the reasons above, we validate our proposed models using PCA.

\subsection{Relationship to Prior Work}

The overwhelming majority of research on CV considered the development of robust algorithms to improve accuracy in static image datasets \cite{mobilefacenets, deepfacedetector} (and references within), 
fewer dealt with videos \cite{kcoveragemobilecam}, and even fewer contributions addressed system design aspects \cite{camstyle}.

In this study, we model the CV accuracy in terms of the main video encoding parameters. None of the prior studies developed accuracy models. Most studies on CV did not even consider the impact of video adaptation on the accuracy.   In   \cite{Sharrab2012},  video adaptation was analyzed in terms of face detection accuracy, without providing any models and without using open datasets.
Although face detection is a simple CV algorithm to implement, it has limited usefulness 
in practice when applied alone. 
Face recognition is much more important as it can precisely reveal subject identity rather than pointing out the presence of an arbitrary subject. Furthermore, deep learning algorithms were not utilized.

Some prior studies on video adaptation considered video quality  metrics, such as  
{\em Mean Squared Error} (MSE) and 
{\em Structural Similarity Index} (SSIM), with  much literature
on rate-distortion optimization \cite{Hsu2011} (and references within). 
In CV systems, however,  the recognition accuracy, not the human perceptual quality, should be the main metric
because the videos are analyzed by machines.

Study \cite{facerecogsurvey} explored the impact of illumination, facial expression, and occlusion on statistical-based face recognition accuracy without any modeling. 
\section{Development of the Proposed Models}
\label{sec:performance}

\subsection{Overview and Motivation}
\label{overview}

We analyze the effectiveness of combining video adaptation strategies in terms of the CV accuracy, focusing primarily on face recognition.
We consider adapting the video streams by changing both the spatial resolution and the Signal-to-Noise Ratio (SNR). We utilize a super-resolution  algorithm  to upscale the videos to their original resolutions  before the analysis at the destinations  in order to boost the accuracy.  
We employ the
Lanczos upscaling algorithm because it  outperforms  other algorithms including 
Bicubic and Spline, in terms of the overall tradeoff between  performance and execution time \cite{Sharrab2012}.
For the SNR adaptation, we consider both changing the target bitrate and $Q_p$.
We do not consider temporal adaptation as missing frames will trivially lead to zero detection and therefore no recognition.

\begin{figure*}[!ht]
	\centering

	\includegraphics[width=\textwidth, trim={10, 35, 15, 15}, clip]{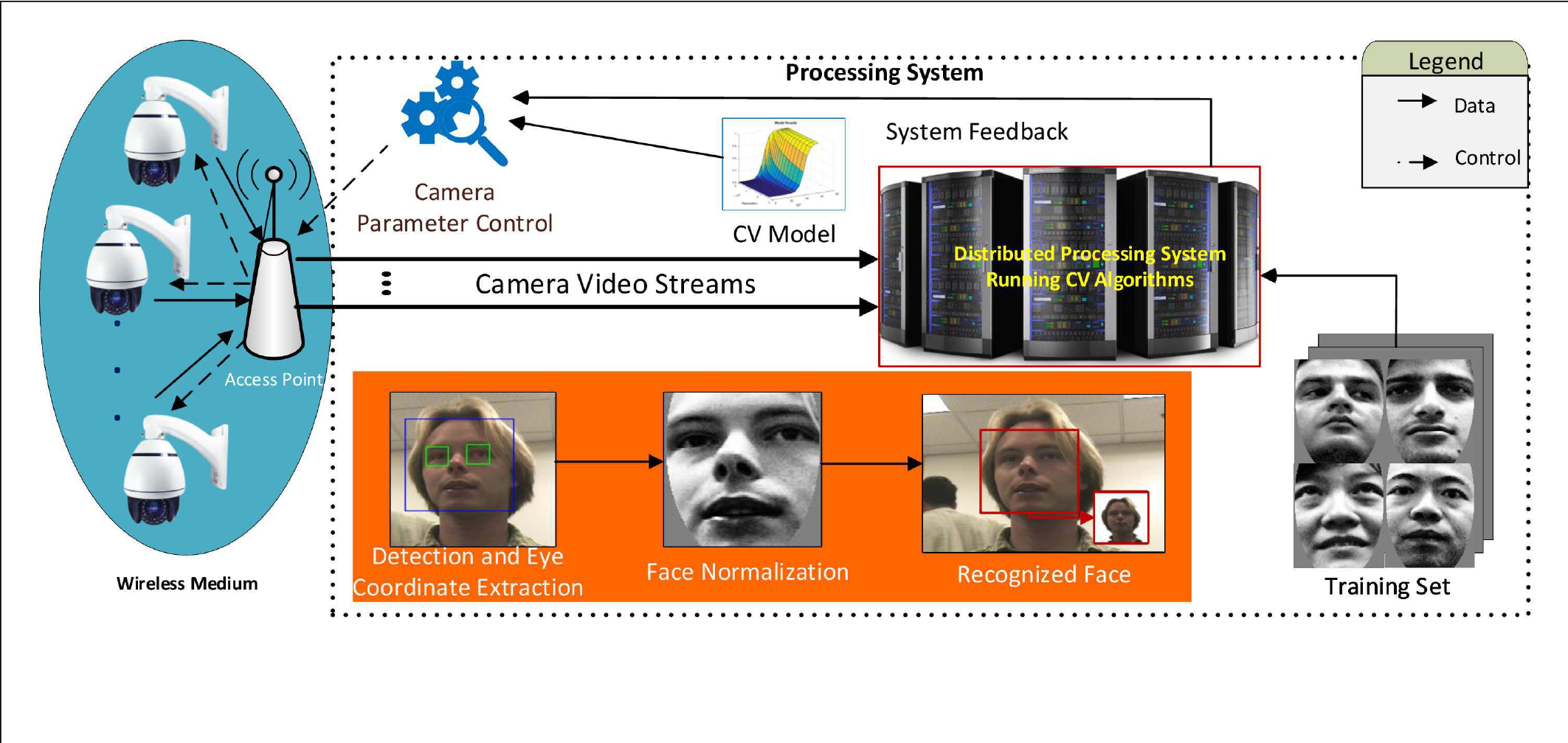}
	\caption{Utilization of the Proposed Models in Controlling the Cameras of a CV System}
	\label{fig:cvsys}

\end{figure*}

Moreover, we  propose two models of the CV accuracy with respect to the parameters used by the aforementioned adaptations.
These models are of great importance and can be utilized to control camera
settings in a way that optimizes the overall CV accuracy.
Figure \ref{fig:cvsys} illustrates a potential use scenario.
In the envisioned system, multiple sources stream video to a distributed processing system for analysis.
to enable the distributed processing system to  optimally determine the settings of each camera
(such as  resolution and target bitrate or quantization parameter).
Eventually, the distributed processing system, running the CV algorithm,
will be able to achieve the optimal recognition accuracy,
given various constraints and the available system and network resources.

\subsection{Model Development}
\label{sec:dev}

We develop two novel models of the CV accuracy in terms of the video encoding parameters. 
The first model characterizes face recognition accuracy with respect to variations in $Q_p$ and resolution, and thus we refer to it as QRMODA ($Q_p$ and Resolution based MODel for Accuracy). 
Similarly, we develop another model for accuracy with respect to the actual bitrate and resolution, and we refer to it as BRMODA (Bitrate and Resolution based MODel for Accuracy). 
Subsequently, we show that both models apply
to face recognition and face detection as well, but with different constant values.  

Though accuracy is the simplest metric to evaluate the performance of any classifier system, other measures such as precision and recall, tell more about the nature of the classifier.
Particularly, these metrics are more important when dealing with imbalanced data. 
When the negative class is dominant, any classifier will more likely predict negative and achieve high accuracy. 
Nonetheless, such a classifier will have no more than $50\%$ precision/recall. This is because the latter two measure the ratio of {\em correctly predicted} to the total {\em positively predicted} and {\em positively actual} classes, respectively. Since recall captures the rate with respect to actual data, rather than predicted, we believe it is more valuable in face recognition applications. In other words, a positively classified (False Positive) face is not a catastrophic issue, whereas an overlooked (false negative) face that should have been flagged as positive, is a major security concern. 
For this reason, we argue that recall is an important measure in face recognition and is more valuable since it captures the sensitivity of the system \cite{recall}.
Recall is also important when it comes to evaluating the performance of a face detector since it can characterize the performance of binary classification tests. An example of such tests is face detection, which can result in either detecting a face or not. 
Many recent literatures also use recall to report system sensitivity, for instance recent literature like \cite{deepfacedetector}, also use recall to measure the performance of face detector adaptation to training with different datasets. 

We use the {\em recall error} (denoted by $\mathcal{E}$) to measure the system sensitivity. Given a video of $k$ frames, $\mathcal{E}$ for the entire video can be given by 

\begin{equation}
 	\mathcal{E} = 1 - \frac {\sum_{i=1}^{k} TP_i}{\sum_{i=1}^{k} TP_i + FN_i},
\end{equation}
where $TP_i$ and $FN_i$ are the numbers of correctly and erroneously identified faces in frame $i$.
Our goal is to characterize $\mathcal{E}$ in terms of simultaneous independently adapting parameters.

\begin{table}
	\begin{center}
		\caption{Used Notations} 
		\label{table:notations}
		\begin{tabular}{|p{0.3\linewidth}|p{0.6\linewidth}|}
			\hline
			Notation& Explanation \\  \hline \hline			
			$\mathcal{E}$&	Recall Error \\ \hline
			$TP$& True Positive \\ \hline
			$FN$& False Negative \\ \hline
			$f_{logistic}(x)$& The Logistic Function \\ \hline
			$Q_p$& Quantization Parameter \\ \hline
			$c_1$, $c_2$, $c_3$, $c_4$, and $c_5$& QRMODA Constants \\ \hline
			$c'_1$, $c'_2$, $c'_3$, $c'_4$, and $c'_5$& BRMODA Constants \\ \hline
			$N\times M$& Video Resolution \\ \hline
			$\mathcal{R}$& Actual Video Bitrate \\ \hline
			$R^2$& Coefficient of Determination \\ \hline					
		\end{tabular}
	\end{center}
\end{table} 

\subsubsection{QRMODA}

Since video adaptation is imposed due to network resource limitations, we expect the CV accuracy to suffer starvation beyond a certain threshold of $Q_p$. 
However, due to a simultaneous independent adaptation in video resolution, a compensation for the accuracy loss will be granted if the video adapts to a higher resolution. Our empirical data, discussed in Section \ref{sec:validation}, indicate that $\mathcal{E}$ follows an exponential trend with respect to changes in resolution and a {\em bounded exponential} bias towards $Q_p$ variations. Hence, we determine that $\mathcal{E}$ is a function that combines the characteristics of both (exponential and bounded exponential) functions, 
which is known as the {\em logistic function}.
We find that $\mathcal{E}$ is a logistic function of $Q_p$ with the x-axis of the Sigmoid's midpoint ($x_0$) being a function of spatial resolution.
Specifically, given a video with a resolution of  $N\times M$, quantized at $Q_p$, $\mathcal{E}$ can be characterized as 

\begin{equation}
\mathcal{E}_{QRMODA} = f_{logistic}(x = Q_p, x_0 = c_1(NM)^{c_2})+c_3,
\label{eq:1}
\end{equation}
where
\begin{equation}
f_{logistic}(x) = \frac{c_4}{1+e^{c_5(x-x_0)}}.
\nonumber 
\end{equation}

\begin{figure}[!ht]
	\centering
	
	\includegraphics[width=0.5\textwidth]{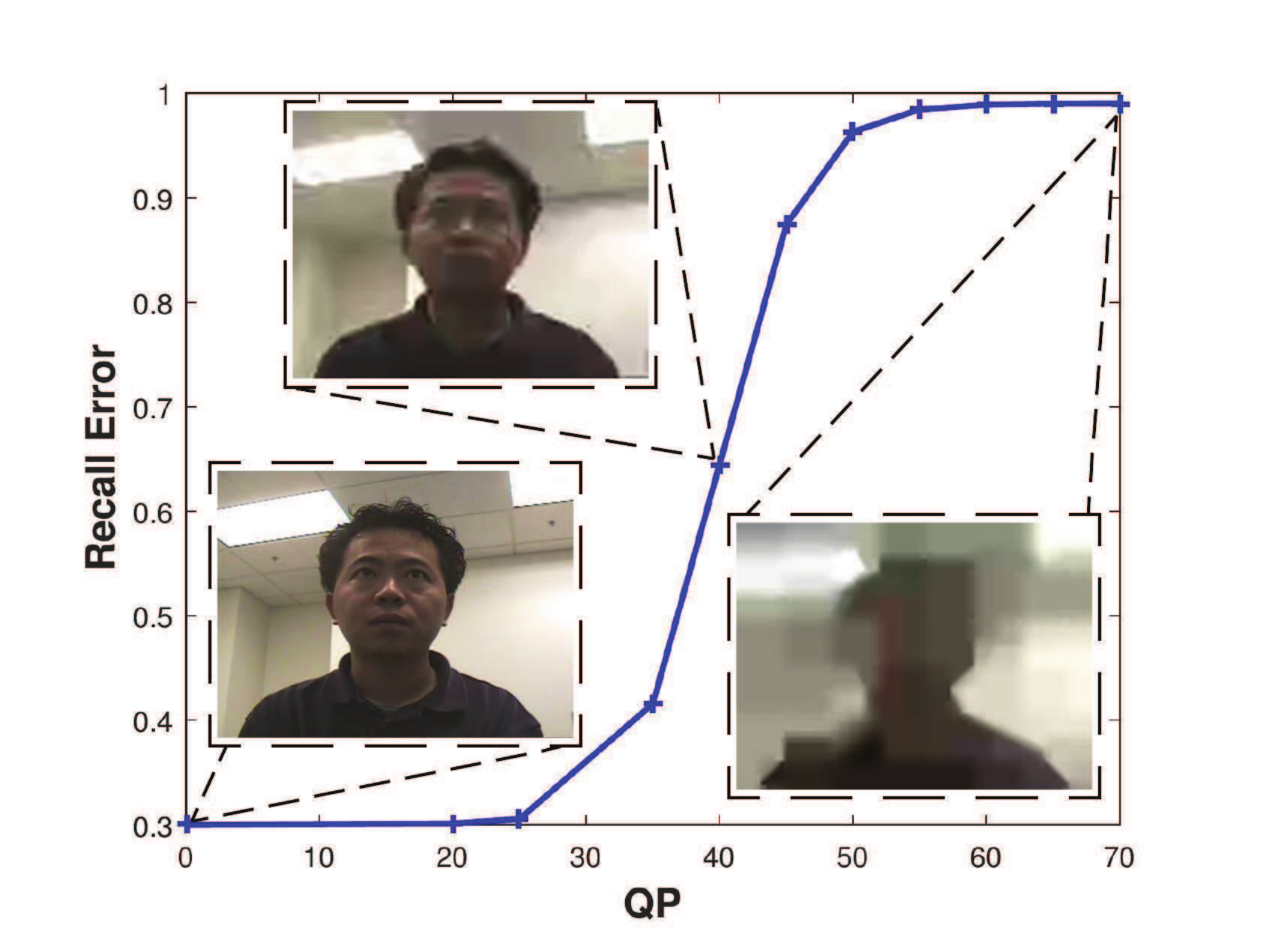}
	\caption{Illustration of the Trend Captured by QRMODA}
	\label{fig:QRMODA}
	
\end{figure}

We introduce $c_3$ as a bias to the model. This value defines the lowest achievable recall error
(i.e. at the original resolution with no quantization).
The model constants $c_1$ through $c_5$ vary based on factors that we will discuss in Section \ref{sec:discussion}.
$c_1$ and $c_2$ define the sharpness in the change of the Sigmoid slope and impact the model's trend with respect to variations in only the spatial resolution. Small values (less than $1$) will result in a smooth transition in recall (with a slope of around $80^{\circ}$, depending on the value of constant $c_5$) that is slightly affected by spatial adaptation. Contrarily, values greater than $1$ will result in a sharp transition in recall as more quantization is imposed (especially with low resolution). As the resolution is increased, the recall transition will flatten. The constant $c_4$ determines the maximum value of the logistic function without the bias. Specifically, ($c_3+c_4$) determine the lowest recall rate, regardless of adaptation variations. Lastly, $c_5$ determines the logistic growth rate (steepness of the curve). Since recall error increases with quantization, 
this is always negative.
Figure \ref{fig:QRMODA} illustrates the  trend captured by the QRMODA model  with sample frames at a fixed resolution but at different $Q_P$ adaptation levels.

\subsubsection{BRMODA}

\begin{figure}[!ht]
	\centering
	
	\includegraphics[width=0.5\textwidth]{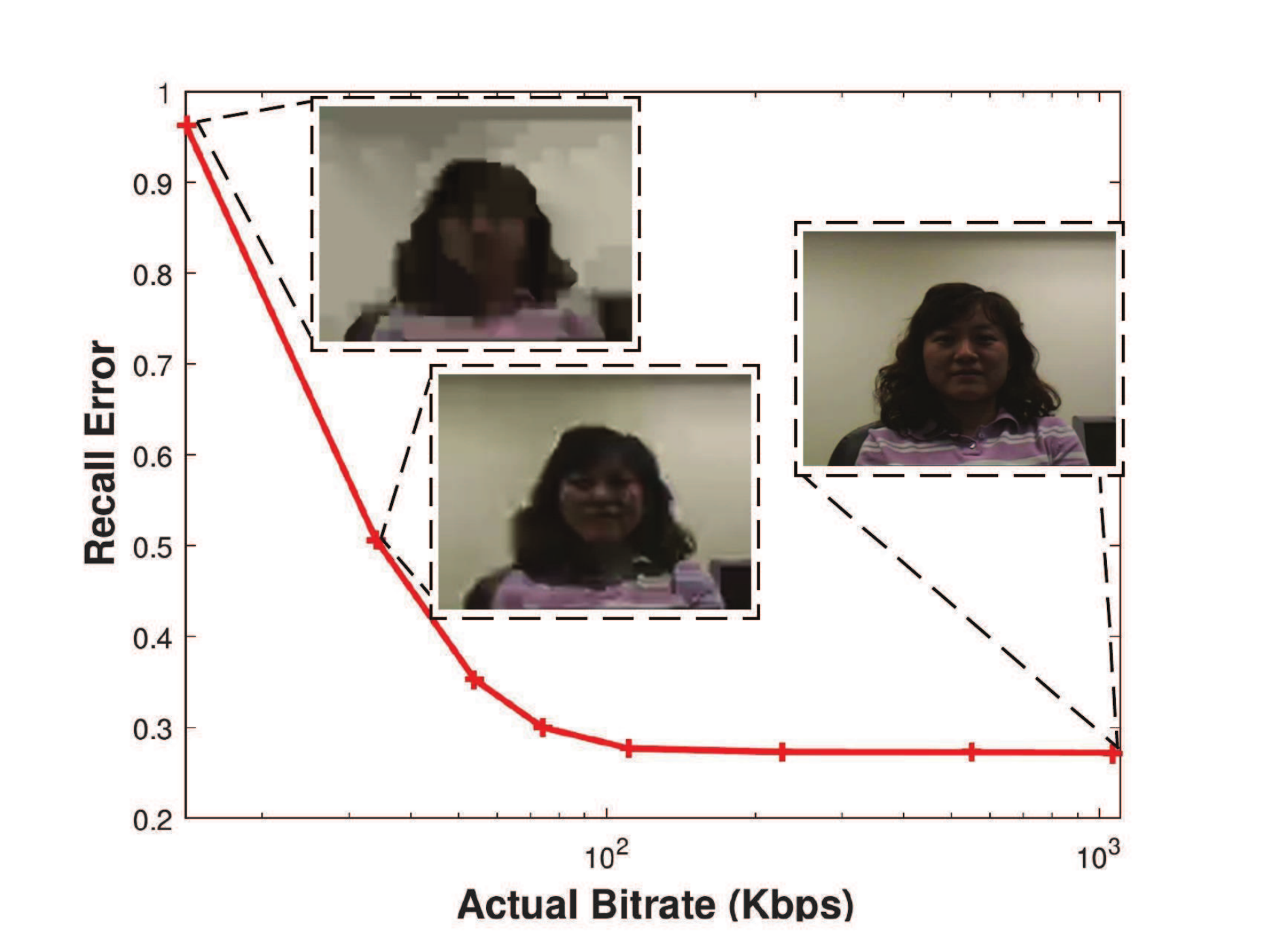}
	\caption{Illustration of the Trend Captured by BRMODA}
	\label{fig:BRMODA}
	
\end{figure}

Videos with low resolutions tend to produce low bitrates when high target bitrates
are imposed. Likewise, videos with high resolutions tend to produce
higher bitrates  than the imposed target values. 
Low bitrate videos have lower recall rates due to reduction in video quality.
As higher bitrates are granted, the video quality increases, thereby causing the recall error to drop drastically. Our empirical results show an exponential relationship. 
We determine that $\mathcal{E}$
is a function of two exponentials of the {\em actual} bitrate, with the number of pixels in the frame being a multiplicative factor with one of the exponentials.
Given an  $N\times M$ resolution video with an actual bitrate $R$,  $\mathcal{E}$ can be given as

\begin{equation}
\mathcal{E}_{BRMODA} = c'_1(NM)^{c'_2}e^{c'_3R}+c'_4e^{c'_5R},
\label{eq:2}
\end{equation}
where $c'_1$ through $c'_5$ are constants.
This model uses the value of the actual bitrate because the target bitrate may not be achieved
precisely by the encoder.  
Constants $c'_1$ and $c'_2$ are similar to their counterparts in QRMODA in terms of purpose. They define the steepness of the exponential drop with respect to spatial resolution variation. 
Constant $c'_3$ is always negative
because $\mathcal{E}$ is inversely proportional to the actual achieved bitrate. In other words, high-resolution videos require high bitrates, and thus will produce high recall errors when low bitrates are imposed.
$c'_4$ and $c'_5$ control the bias exponential.
Figure \ref{fig:BRMODA} shows the trend captured by the BRMODA model, with sample frames at a fixed resolution but with different bitrates. 

\section{Experimental Setup}
\label{sec:setup}

\subsection{Used Datasets}
\label{subsec:datasets}

We utilize two greatly distinct video datasets: Honda/UCSD, and DISFA.
The former is a standard
video database provided for the evaluation of face detection,
tracking, and recognition algorithms. The latter is used to study
{\em Facial Action Coding Systems} (FACS). 
Honda/UCSD has lower  quality videos, which serve as
an example of limited-bandwidth network systems.
In contrast, DISFA has High Definition (HD)  quality videos. In addition,
the subjects in Honda/UCSD make
different combinations of 2-D and 3-D head rotations and have different facial expressions with varying speed.
On the other hand, subjects in DISFA have limited pose
variations, but great variations in facial action expressions.

Furthermore, we utilize a large image dataset: LFW \cite{lfw}. This dataset aims at studying the problem of unconstrained face recognition.
The main properties of all the used datasets are summarized in Table \ref{table:datasets}.
We divide each  database into three main sets:
\textit{Training}, \textit{Validation}, and \textit{Testing}. 
In Honda/UCSD, we use the first included dataset, which is already categorized into $3$ groups: Training, Testing, and Testing with Partial Occlusion. We use the latter for validation.
Contrarily in DISFA, we split the right camera videos to training and validation sets and use the left camera videos for testing. For LFW, we use the split method suggested by \cite{lfw}. 
The adaptation is performed only on  the Testing sets to avoid overfitting and selection bias towards adapted frames.

\subsection{Video Adaptation Generation}
\label{subsec:adaptation}

We perform H.264 encoding on the testing set videos/images of all datasets using FFmpeg to achieve
different adaptation levels. 
We generate two sets of doubly adapted videos to analyze the impact of two encoding parameters on CV accuracy.
The first set includes videos with combined resolution and target bitrate adaptations, whereas the second set includes videos that have a combination of $Q_p$ and resolution adaptations.
We use the Lanczos algorithm to upscale the videos, 
as it provides the best tradeoff in performance and execution time \cite{Sharrab2012}.
We generate an additional set of doubly adapted images using the testing image set of the LFW dataset. For this set, only a combination of $Q_p$ and resolution is used because bitrate adaptation is inapplicable to images.

\begin{table*}[!ht]
	\begin{center}
		\caption{Characteristics of the Used Datasets} 
		\label{table:datasets}
			\begin{tabular}{|p{0.2\linewidth}|p{0.25\linewidth}|p{0.25\linewidth}|p{0.21\linewidth}|}
				\hline
				Characteristic& Honda/UCSD& DISFA& LFW \\  \hline \hline			
				Camera&	SONY EVI-D30& PtGrey stereo& Varies	\\ \hline
				Subjects&	$20$ ($2$ females and $18$ males)& $27$ ($12$ females and $15$ males)& $5,749$	\\ \hline
				Resolution&	$640\times 480$& $1024\times 768$& $250\times 250$	\\ \hline
				Frame Rate&	$15$ frame/sec& $20$ frame/sec& N/A	\\ \hline	
				Format&	Uncompressed AVI& Uncompressed AVI& JPEG	\\ \hline
				Size& $45$ videos& $54$ videos& $13,233$ images \\ \hline						
			\end{tabular}
	\end{center}
\end{table*}

\subsection {Face Detection and Recognition Implementations}
\label{subsec:neuralrecog}

We use {\bf CNN-based} face detection and recognition,
utilizing FaceNet \cite{facenet} as the deep learning platform.
We develop an interface that interacts with FaceNet to perform normalization, CNN training, face detection, and face recognition. 
The experiments start by organizing all training frames in a tree like fashion such that each subject 
maintains its own directory of the respective frames. These frames are then aligned, maintaining the same
directory structure. The aligned frames are then used to train the fully 
connected layers of the deep CNN, generating a classifier model file for use by the recognition module. 
Subsequently, we fine tune this model using the validation videos.

We use the testing set videos 
as an input to the CNN, and detect the faces in every frame of those videos using FaceNet. We employ the aforementioned classifier model
to classify each frame. The result of this step is a list of probabilities for each probe with respective classes.
We pick the class with the highest probability and consider it as the best candidate identifying the probe (Top 1 class). 
Finally, we collect a {\em confusion matrix}, which we use to compute the overall recall, precision, and F1-score of each experiment.

In the {\bf statistical-based} approach, we develop a face detector using the Viola-Jones \cite{Viola} face detection algorithm that is implemented in OpenCV.  
We develop a platform for extracting eye coordinates from all frames.
Since eye classifiers are not accurate and may return falsely detected eyes, we develop a mechanism to filter true eyes based on their sagittal coordinates. 
These coordinates are vitally important for recognition because they represent input parameters for the preprocessing steps, including  geometric normalization, histogram equalization, and masking.
We utilize the {\em CSU Face Identification Evaluation System} \cite{csu} to perform training and face recognition.
We employ PCA because of its effectiveness in generating  simpler representations of the huge video dataset with all adaptations.
\section{Model Validation and Analysis}
\label{sec:validation}

\subsection{Baselines and Evaluation Metrics}
\label{subsec:baseline}

We use two baselines to benchmark the validity of BRMODA and QRMODA. 
These baselines represent deep learning FaceNet (NN2 architecture \cite{facenet}) and statistical (PCA) face recognition methods. We also employ another baseline for face detection using Viola-Jones algorithm.
Although the methods used by \cite{facenet} and \cite{csu} perform image analysis, we develop interfaces to work with adapted video frames from Honda/UCSD and DISFA datasets.
We use $R^2$ to assess the goodness of fit of the proposed models, and our accuracy metrics are recall, precision, and F1-score. The $R^2$ values are shown with each figure subcaption.

\begin{table*}[!ht]
	\begin{center}
		\caption{List of Constants for CNN-based QRMODA/BRMODA [Detection, Recognition]} 
		\label{table:QRMODACNN}
		\begin{tabular}{|l|r|r|}
			\hline
			Const.& Honda/UCSD& DISFA \\  \hline \hline
			$c_1$&	$17.98, 24.03$& $0.7, 1.54$	\\ \hline
			$c_2$&	$0.08493, 0.05211$& $1.255, 1.121$	\\ \hline
			$c_3$&	$0.5, 0.61$& $0.003, 0.003$	\\ \hline
			$c_4$&	$0.5, 0.3838$& $0.039, 0.5913$	\\ \hline
			$c_5$&	$-0.2, -0.2864$& $-0.4, -0.517$	\\	
			\hline		
			$c'_1$& $0.414, 0.0363$& $2.64\times10^{-4}, 1.867\times10^{-6}$	\\ \hline
			$c'_2$& $0.175, 0.292$& $0.65, 1.02$	\\ \hline
			$c'_3$& $-0.126, -0.054$& $-0.2, -0.117$	\\ \hline
			$c'_4$& $0.174, 0.273$& $0.0229, 0.06102$	\\ \hline
			$c'_5$& $-7.97\times10^{-6}$, $-4.718\times10^{-6}$& $-4.8\times10^{-6}$, $-3.03\times10^{-6}$	\\ \hline	
		\end{tabular}
	\end{center}
\end{table*}

\begin{figure*}[!ht]
	\centering
	
	\subfigure[$600\times 450$ ($R^2$: 0.992, 0.995) Honda/UCSD]
	{
		\minipage{0.23\textwidth}
		\label{fig:neuralqp600}
		\includegraphics[width=\linewidth, trim={15 0 30 0}, clip]{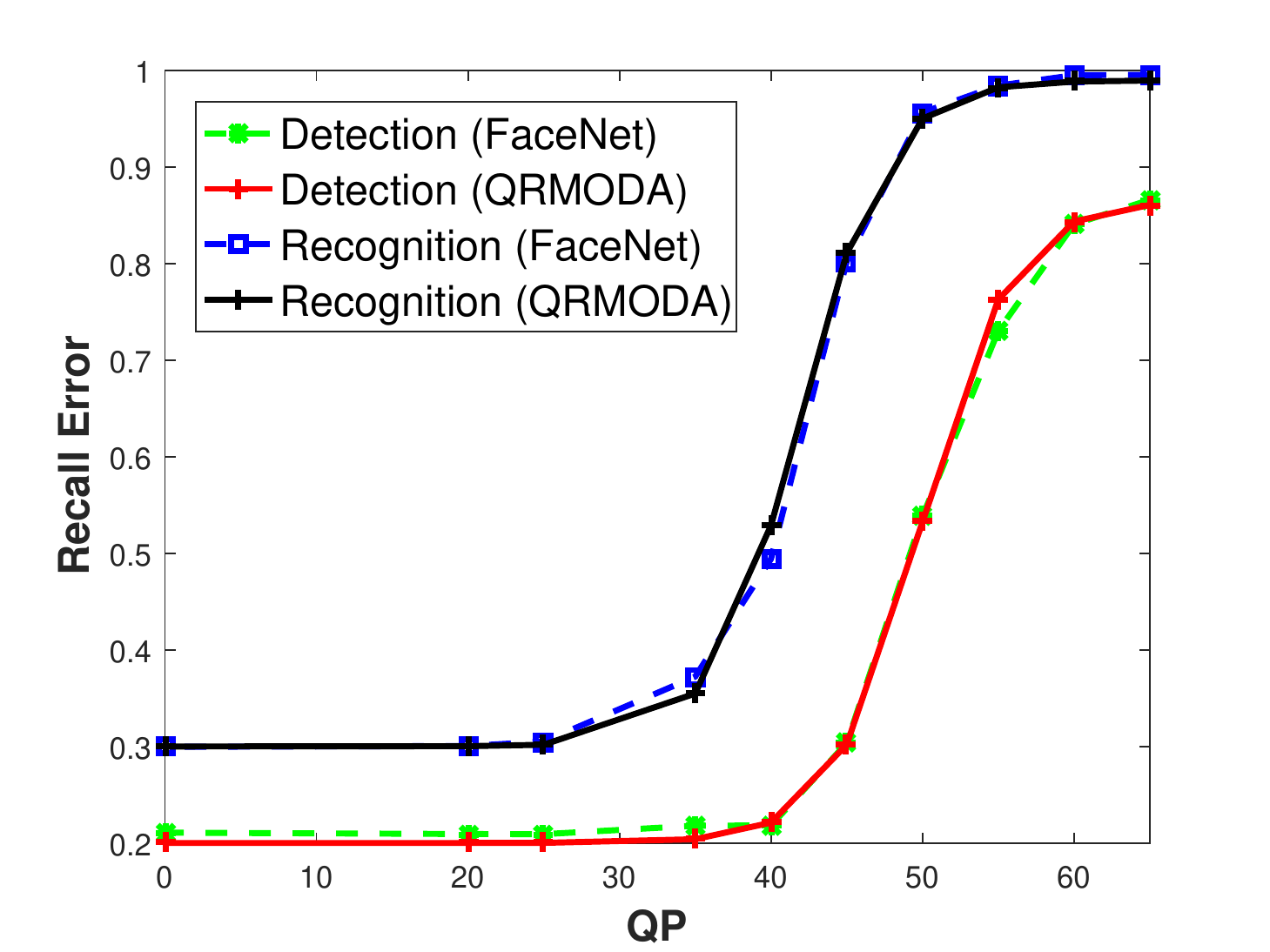}
		\endminipage\hfill
	}
	\subfigure[$520\times 390$ ($R^2$: 0.989, 0.993) Honda/UCSD]
	{
		\minipage{0.23\textwidth}
		\label{fig:neuralqp520}
		\includegraphics[width=\linewidth, trim={15 0 30 0}, clip]{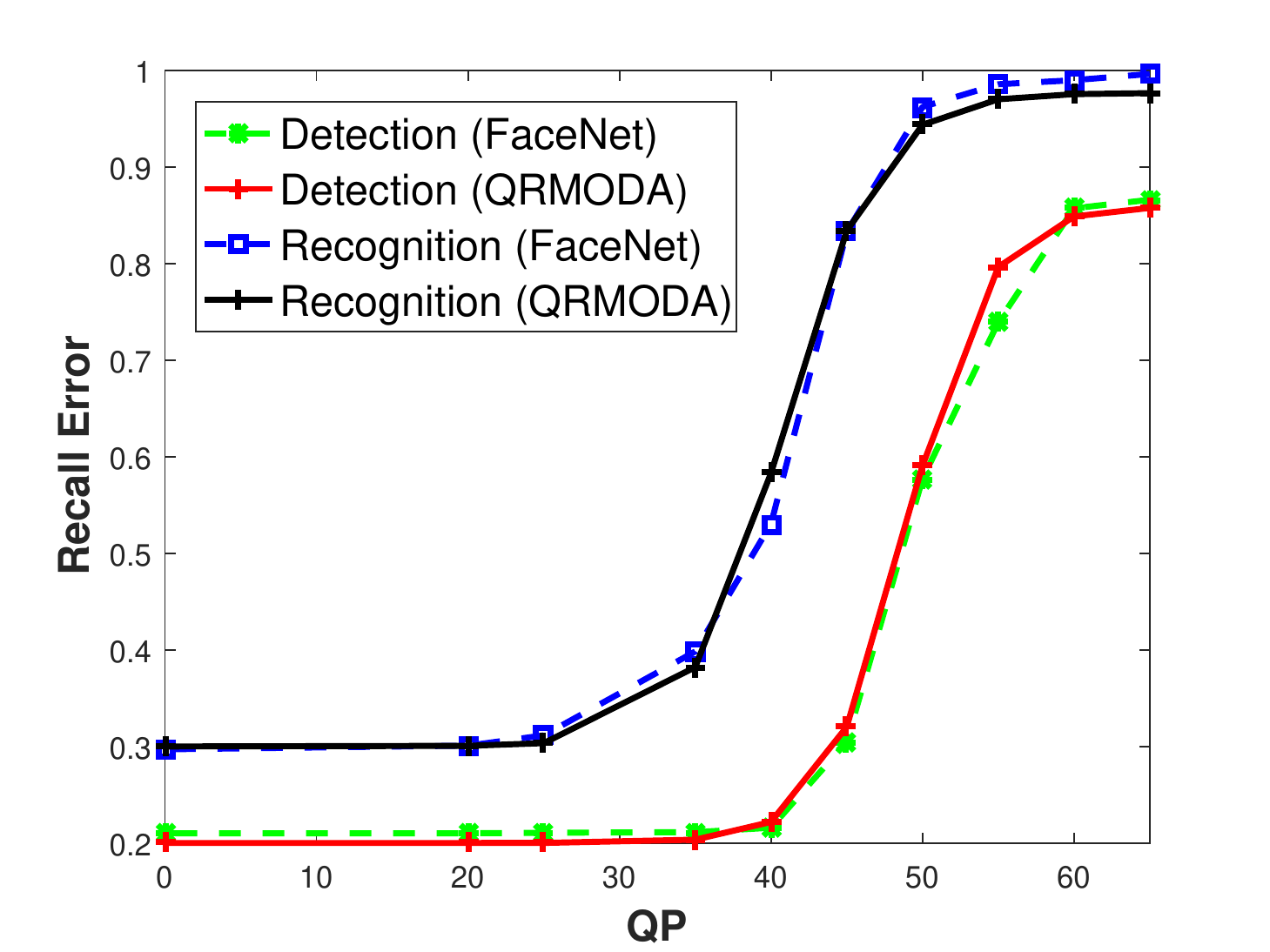}
		\endminipage\hfill
	}
	\subfigure[$400\times 300$ ($R^2$: 0.988,0.997) Honda/UCSD]
	{
		\minipage{0.23\textwidth}
		\label{fig:neuralqp400}
		\includegraphics[width=\linewidth, trim={15 0 30 0}, clip]{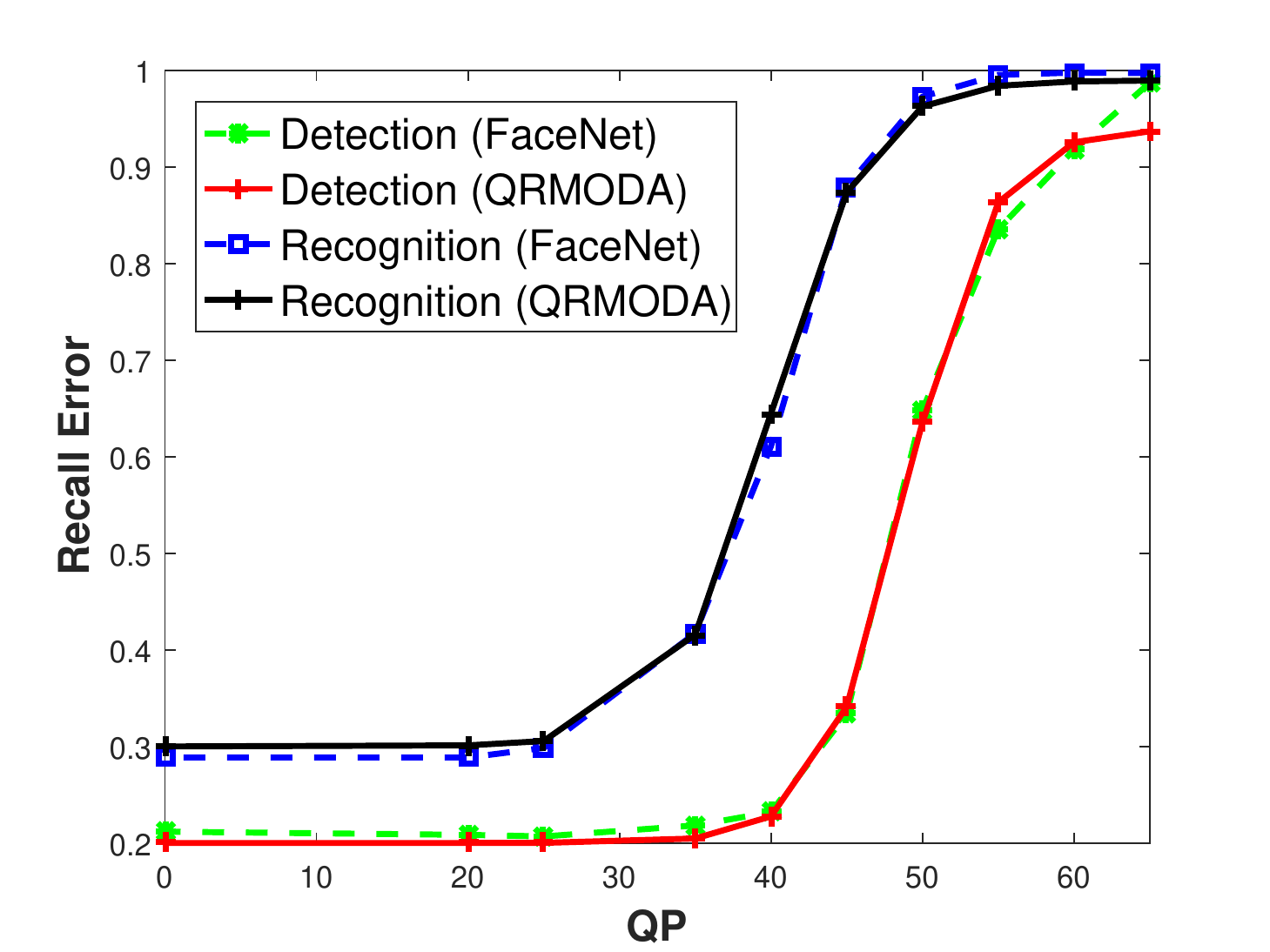}
		\endminipage\hfill
	}
	\subfigure[$280\times 210$ ($R^2$: 0.982, 0.998) Honda/UCSD]
	{
		\minipage{0.23\textwidth}
		\label{fig:neuralqp280}
		\includegraphics[width=\linewidth, trim={15 0 30 0}, clip]{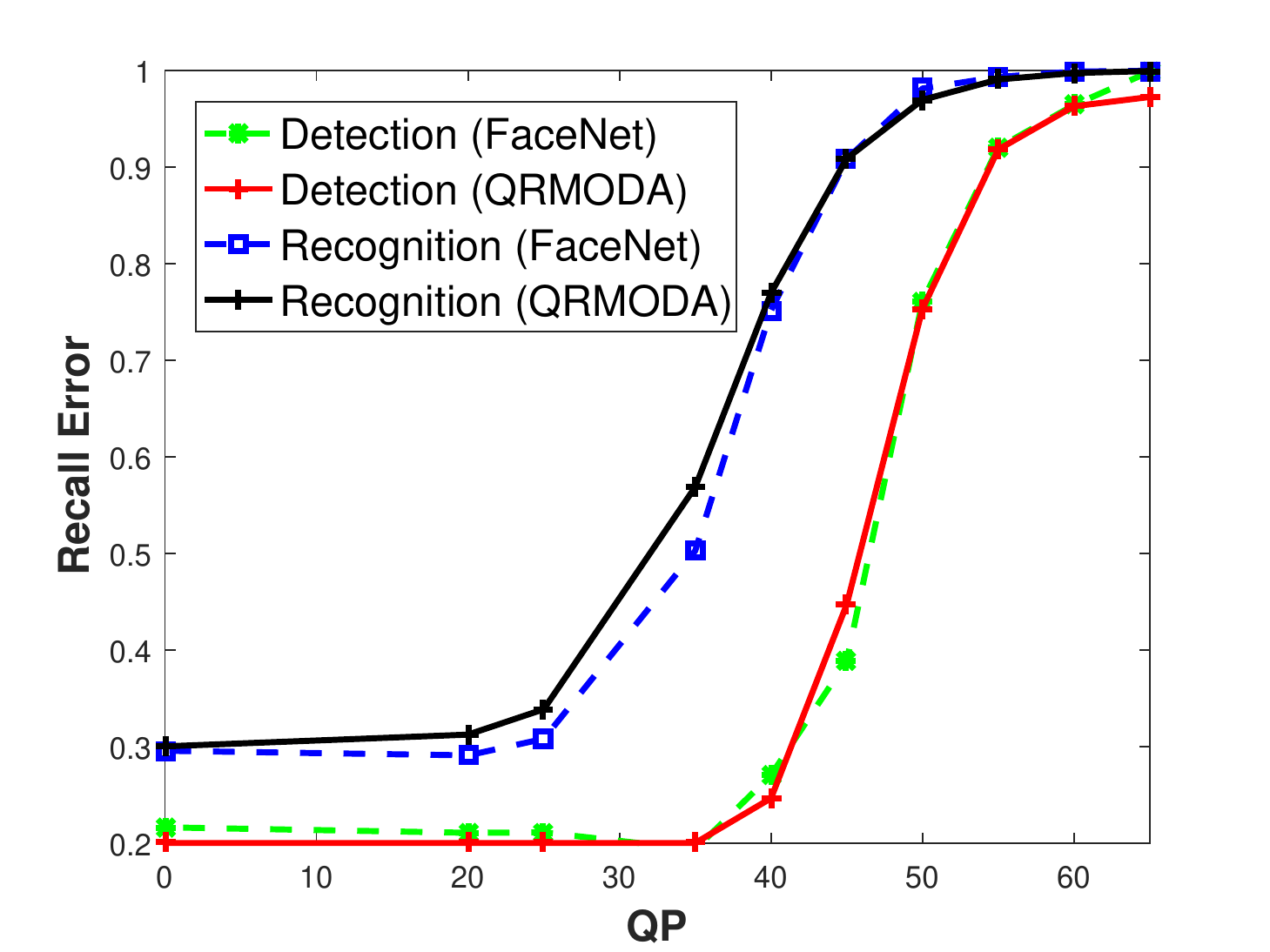}
		\endminipage\hfill
	}
	\subfigure[$600\times 450$ ($R^2$: 0.9962, 0.999) Honda/UCSD]
	{
		\minipage{0.23\textwidth}
		\label{fig:error600}
		\includegraphics[width=\linewidth, trim={15 0 30 0}, clip]{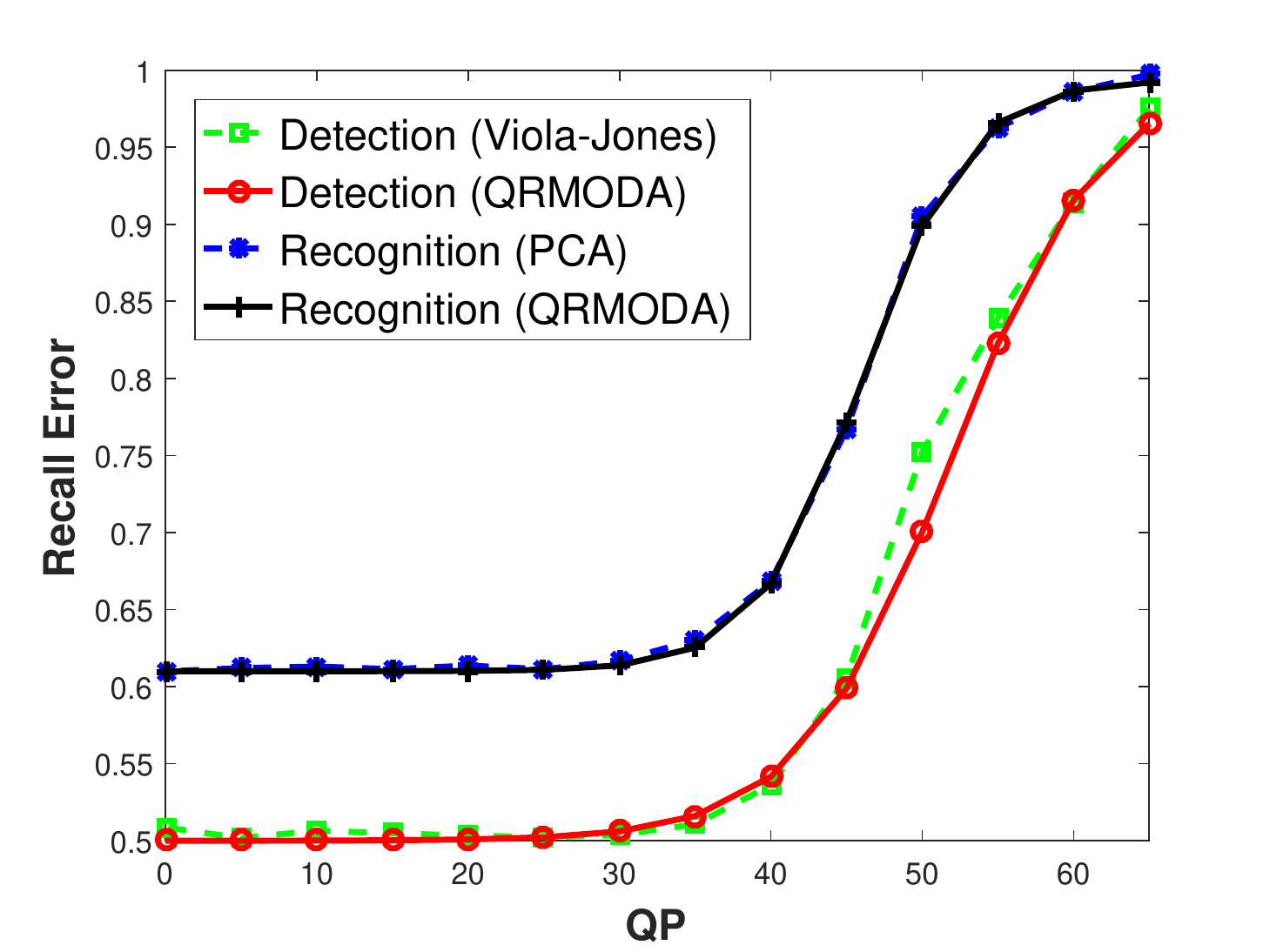}
		\endminipage\hfill
	}
	\subfigure[$520\times 390$  ($R^2$: 0.9949, 0.0.997) Honda/UCSD]
	{
		\minipage{0.23\textwidth}
		\label{fig:error520}
		\includegraphics[width=\linewidth, trim={15 0 30 0}, clip]{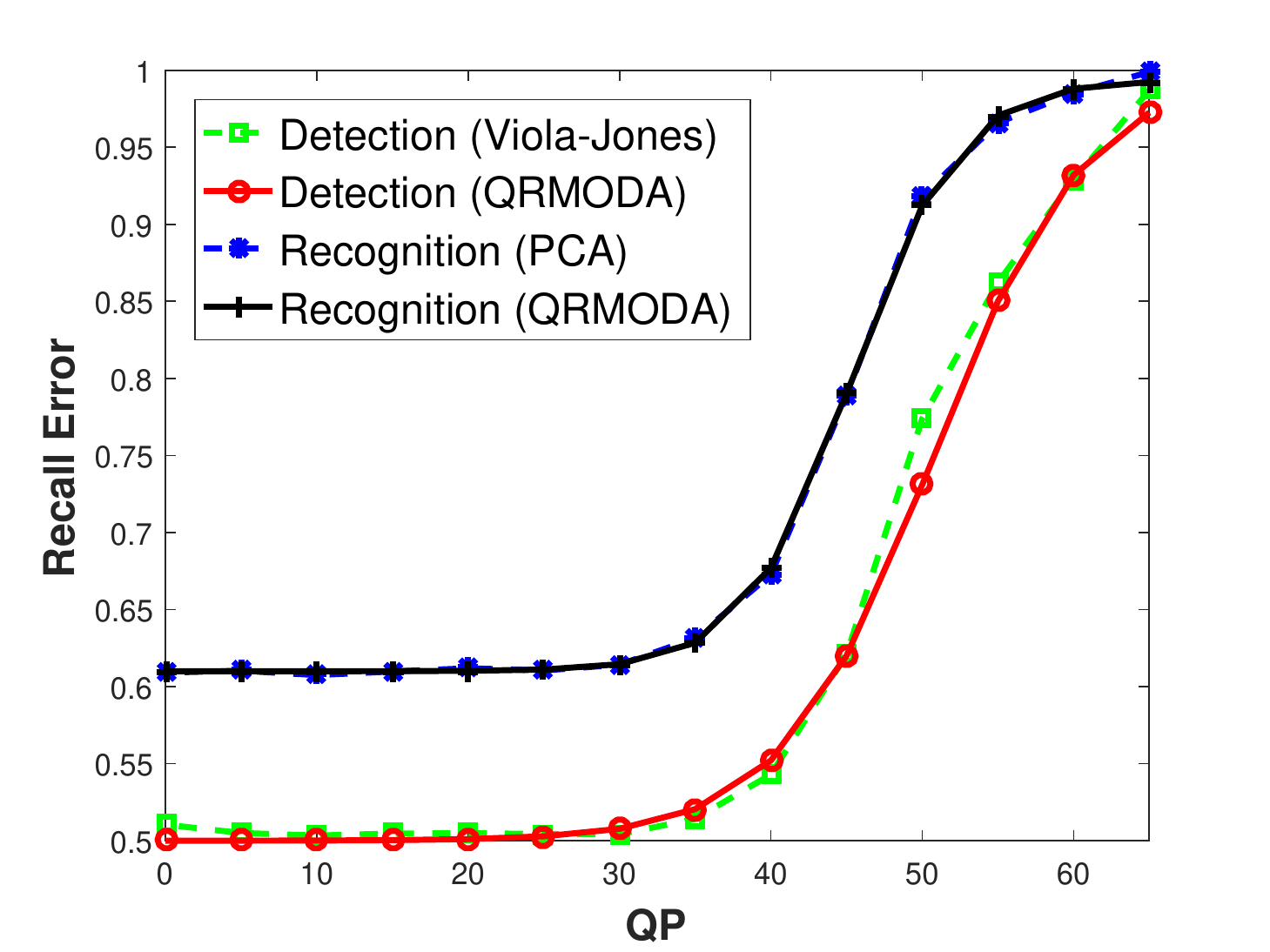}
		\endminipage\hfill
	}
	\subfigure[$400\times 300$  ($R^2$: 0.993, 0.989) Honda/UCSD]
	{
		\minipage{0.23\textwidth}
		\label{fig:error400}
		\includegraphics[width=\linewidth, trim={15 0 30 0}, clip]{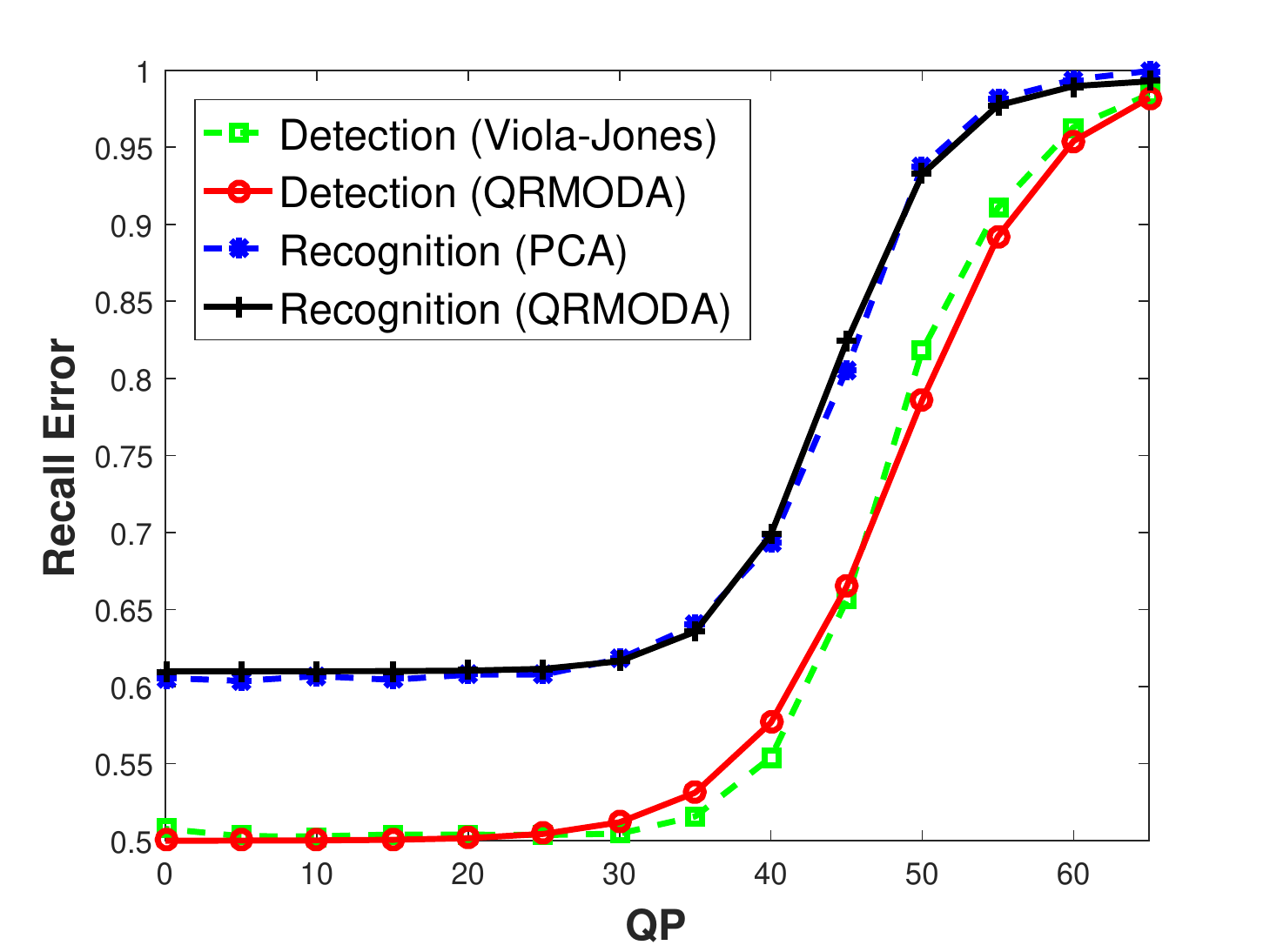}
		\endminipage\hfill
	}
	\subfigure[$240\times 180$  ($R^2$: 0.973, 0.962) Honda/UCSD]
	{
		\minipage{0.23\textwidth}
		\label{fig:error240}
		\includegraphics[width=\linewidth, trim={15 0 30 0}, clip]{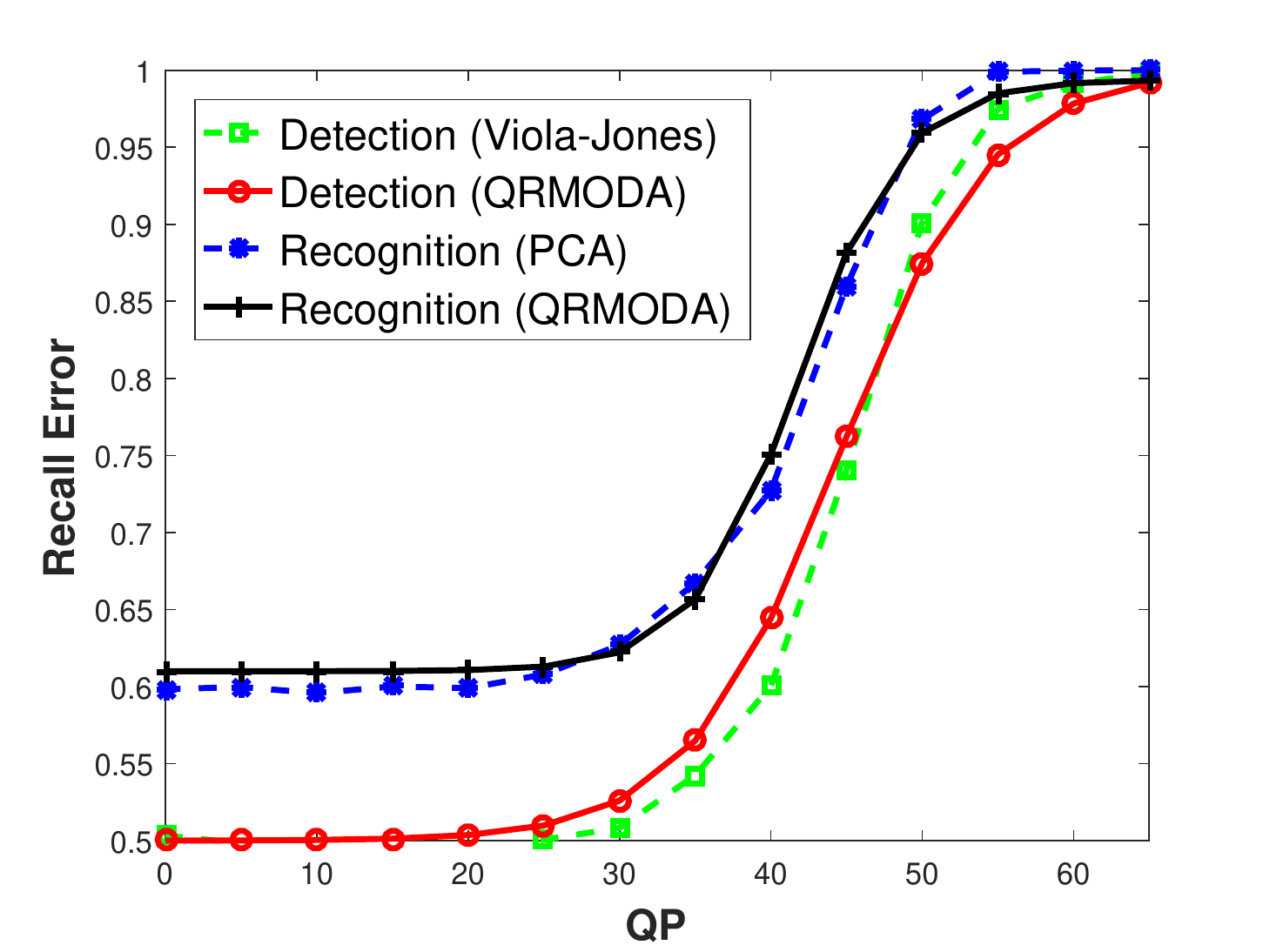}
		\endminipage\hfill
	}
	\subfigure[$480\times 360$  ($R^2$: 0.9968, 0.998) DISFA]
	{
		\minipage{0.23\textwidth}
		\label{fig:DISFAqp480}
		\includegraphics[width=\linewidth, trim={15 0 30 0}, clip]{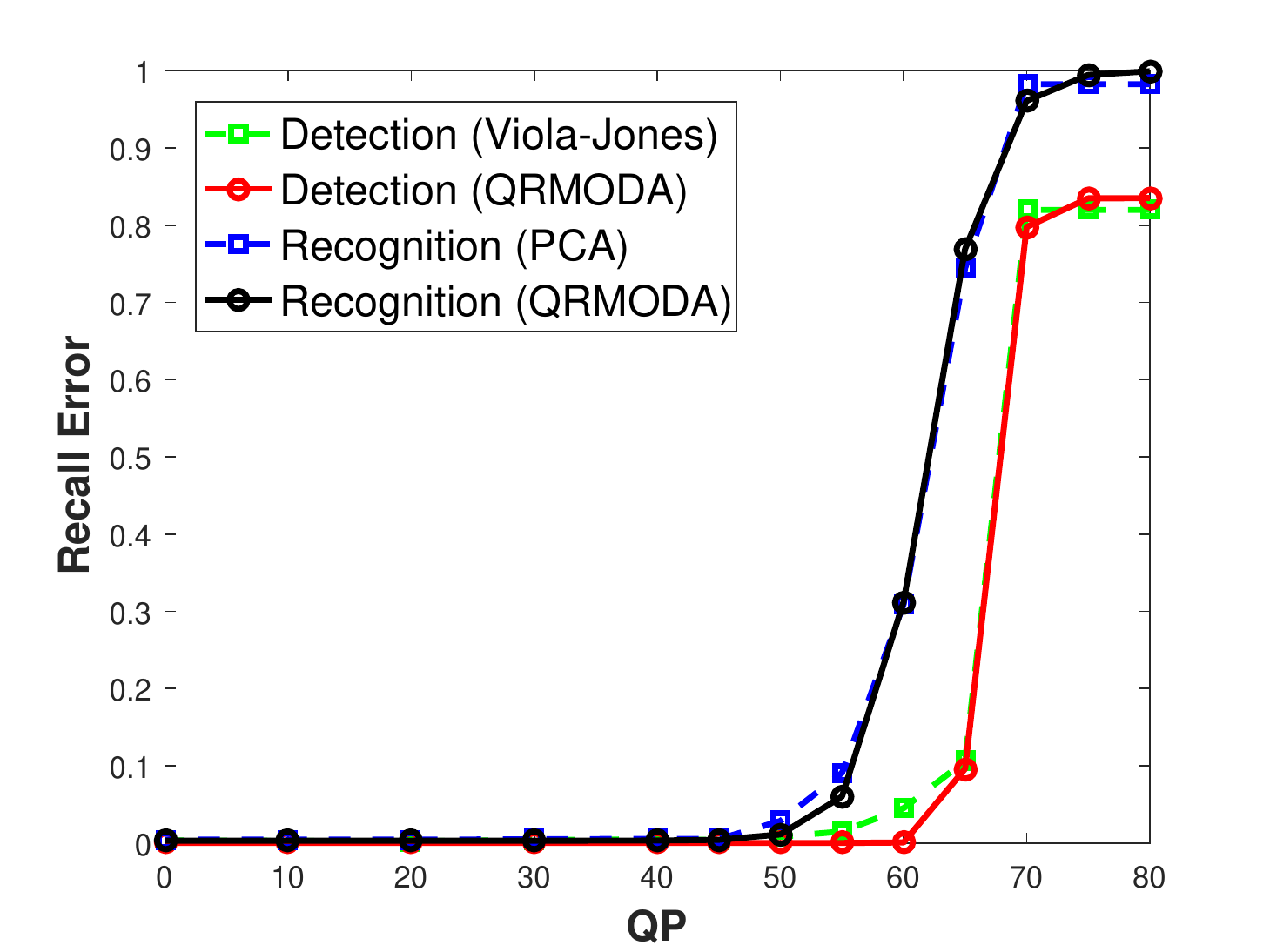}
		\endminipage\hfill
	}
	\subfigure[$280\times 210$  ($R^2$: 0.895, 1) DISFA]
	{
		\minipage{0.23\textwidth}
		\label{fig:DISFAqp280}
		\includegraphics[width=\linewidth, trim={15 0 30 0}, clip]{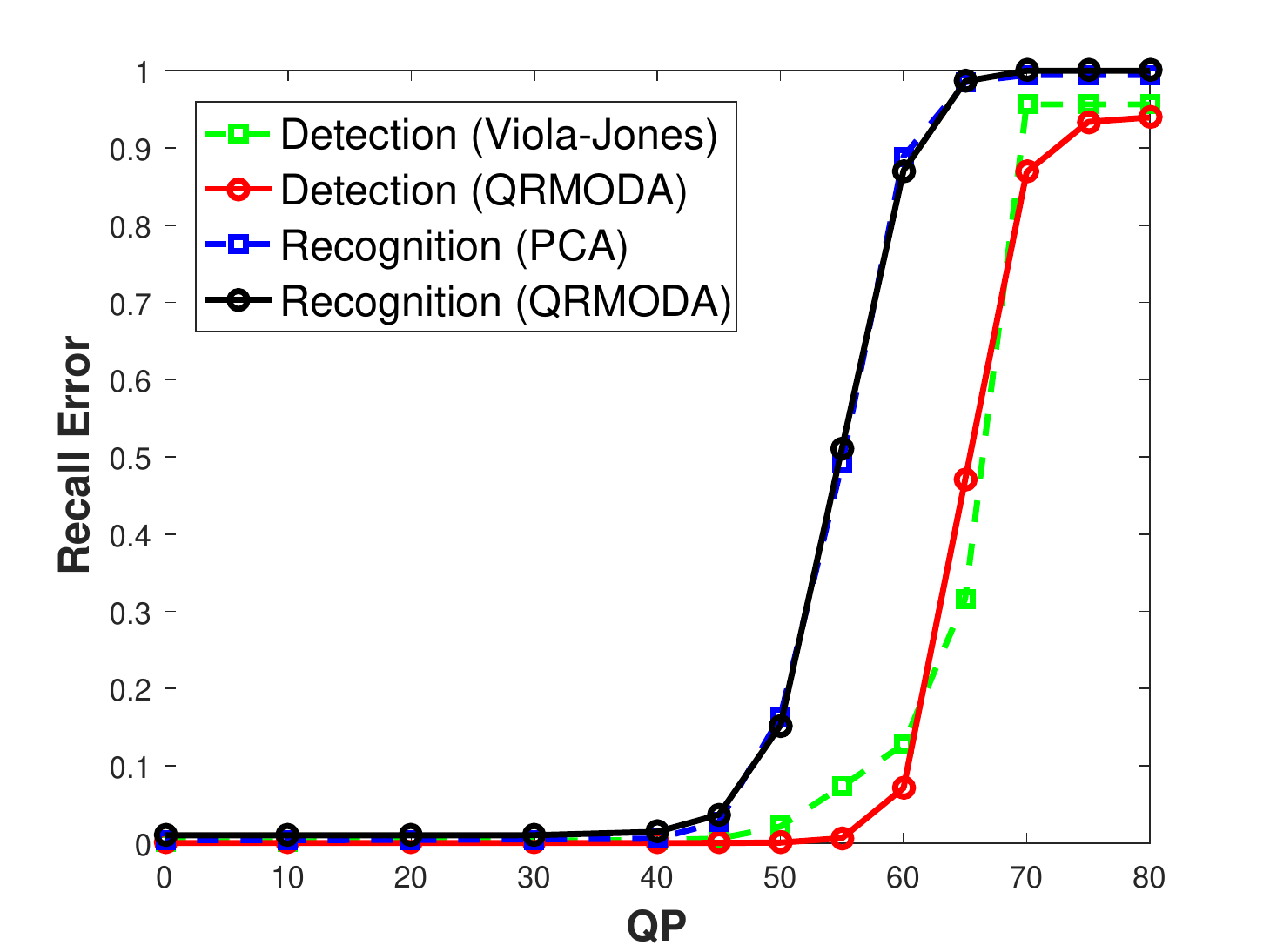}
		\endminipage\hfill
	}
	\subfigure[$50\times 50$  ($R^2$: 0.998) LFW]
	{
		\minipage{0.23\textwidth}
		\label{fig:lfw50}
		\includegraphics[width=\linewidth, trim={15 0 30 0}, clip]{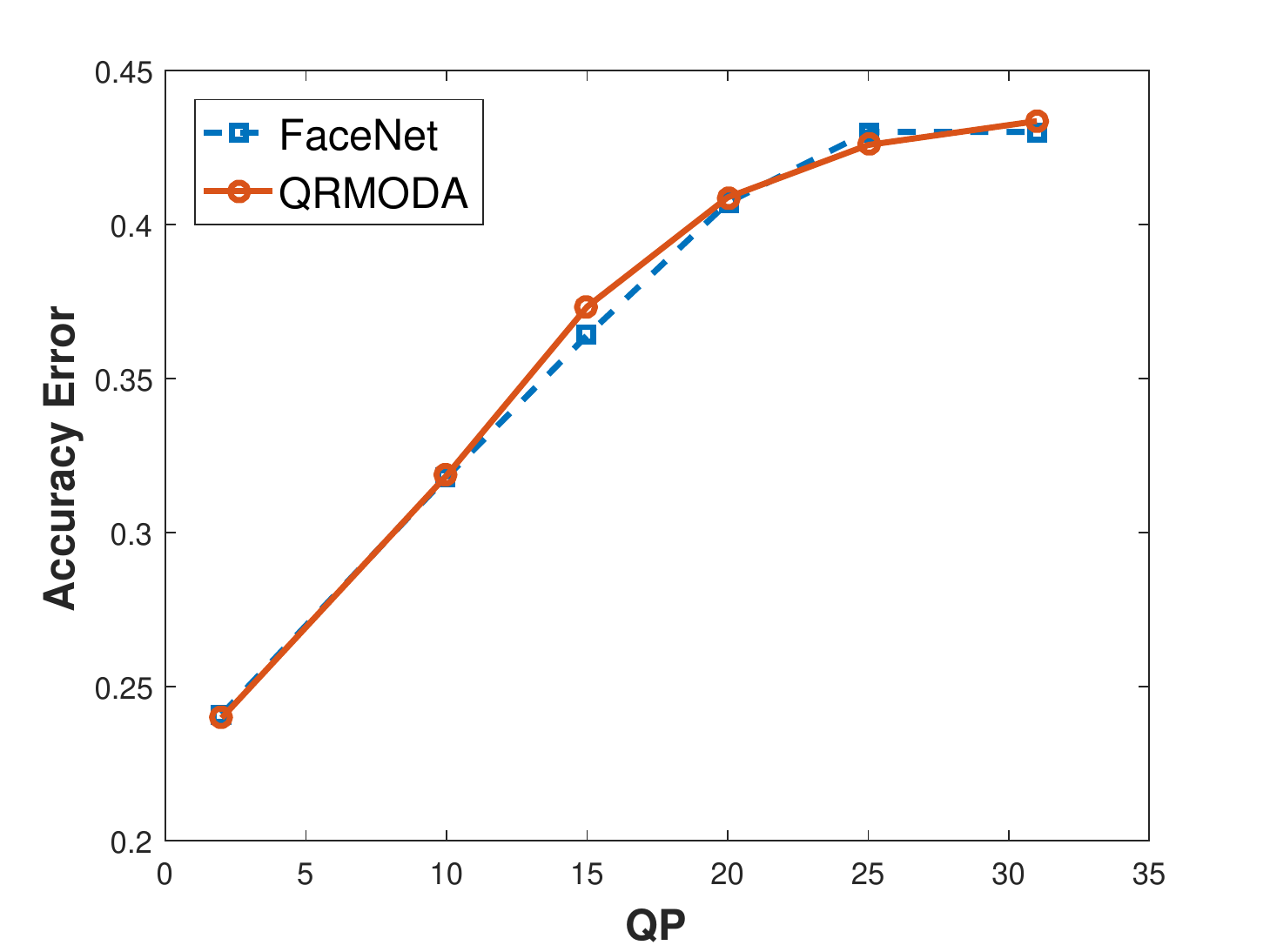}
		\endminipage\hfill
	}
	\subfigure[$200\times 200$  ($R^2$: 0.992) LFW]
	{
		\minipage{0.23\textwidth}
		\label{fig:lfw200}
		\includegraphics[width=\linewidth, trim={15 0 30 0}, clip]{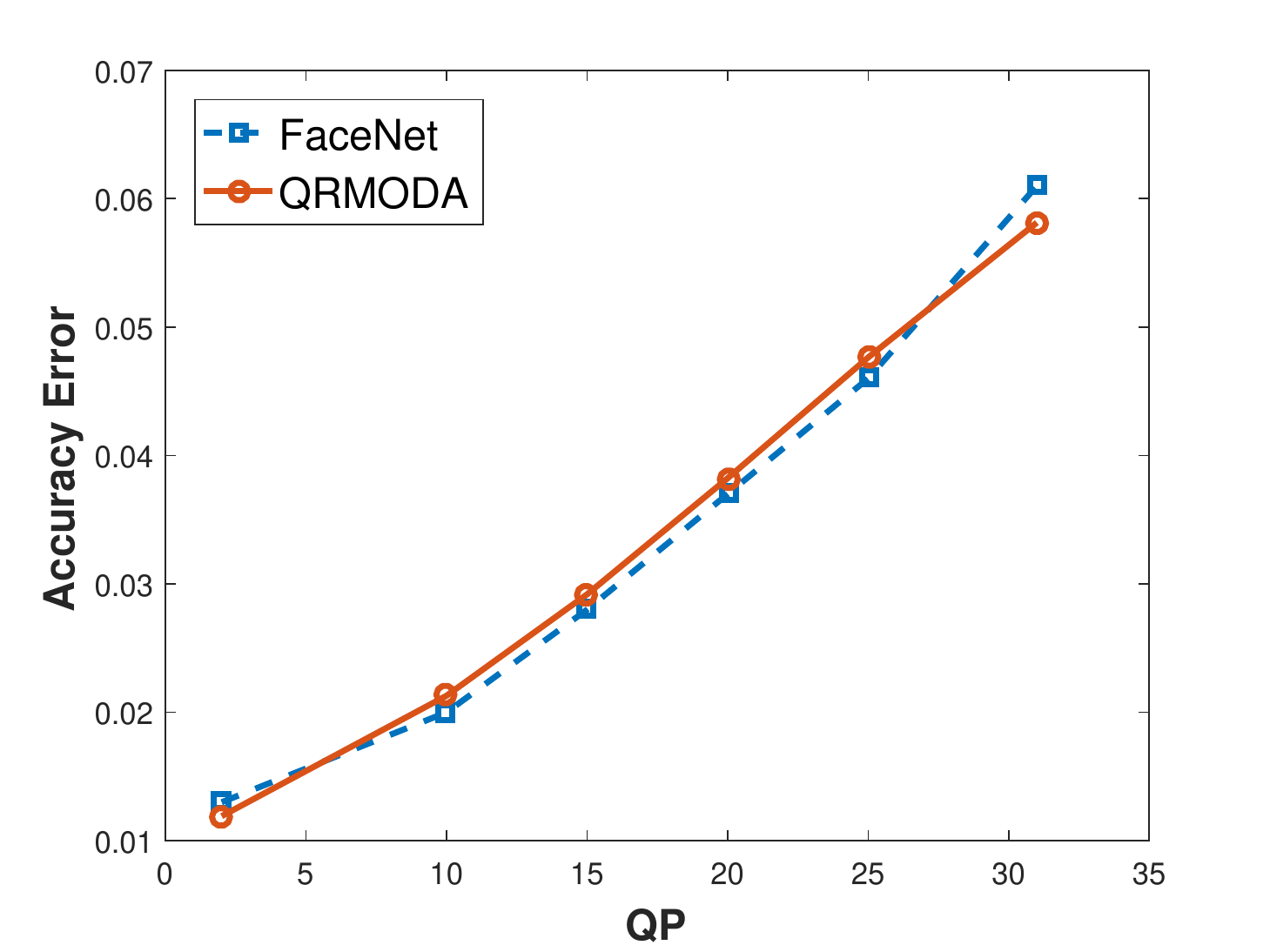}
		\endminipage\hfill
	}
	\subfigure[PCA (HONDA/UCSD)]
	{
		\minipage{0.23\textwidth}
		\label{fig:error3dqp_experimental}
		\includegraphics[width=\linewidth, trim={15 0 30 0}, clip]{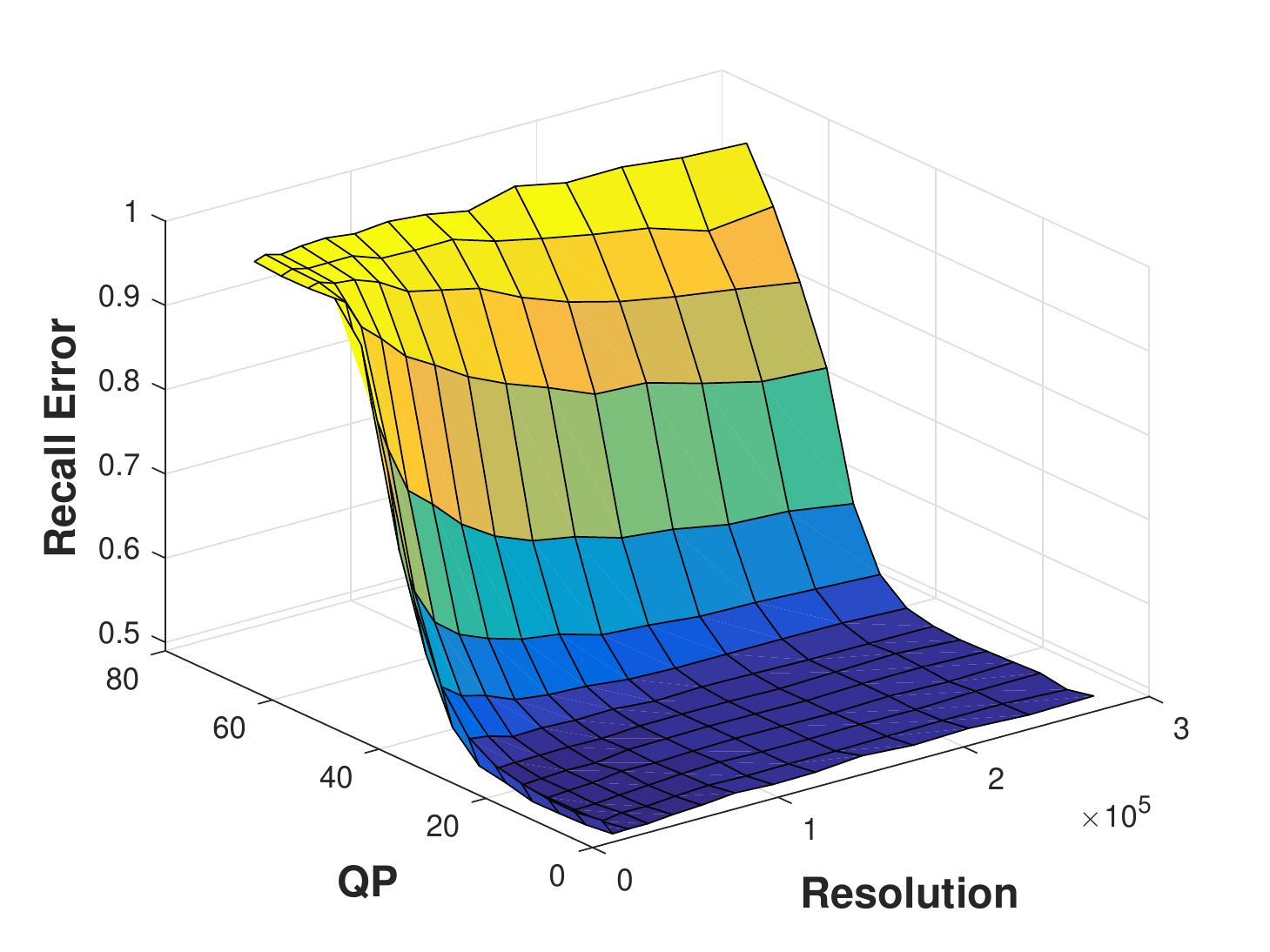}
		\endminipage\hfill
	}
	\subfigure[QRMODA (HONDA/UCSD)]
	{
		\minipage{0.23\textwidth}
		\label{fig:error3dqp_model}
		\includegraphics[width=\linewidth, trim={15 0 30 0}, clip]{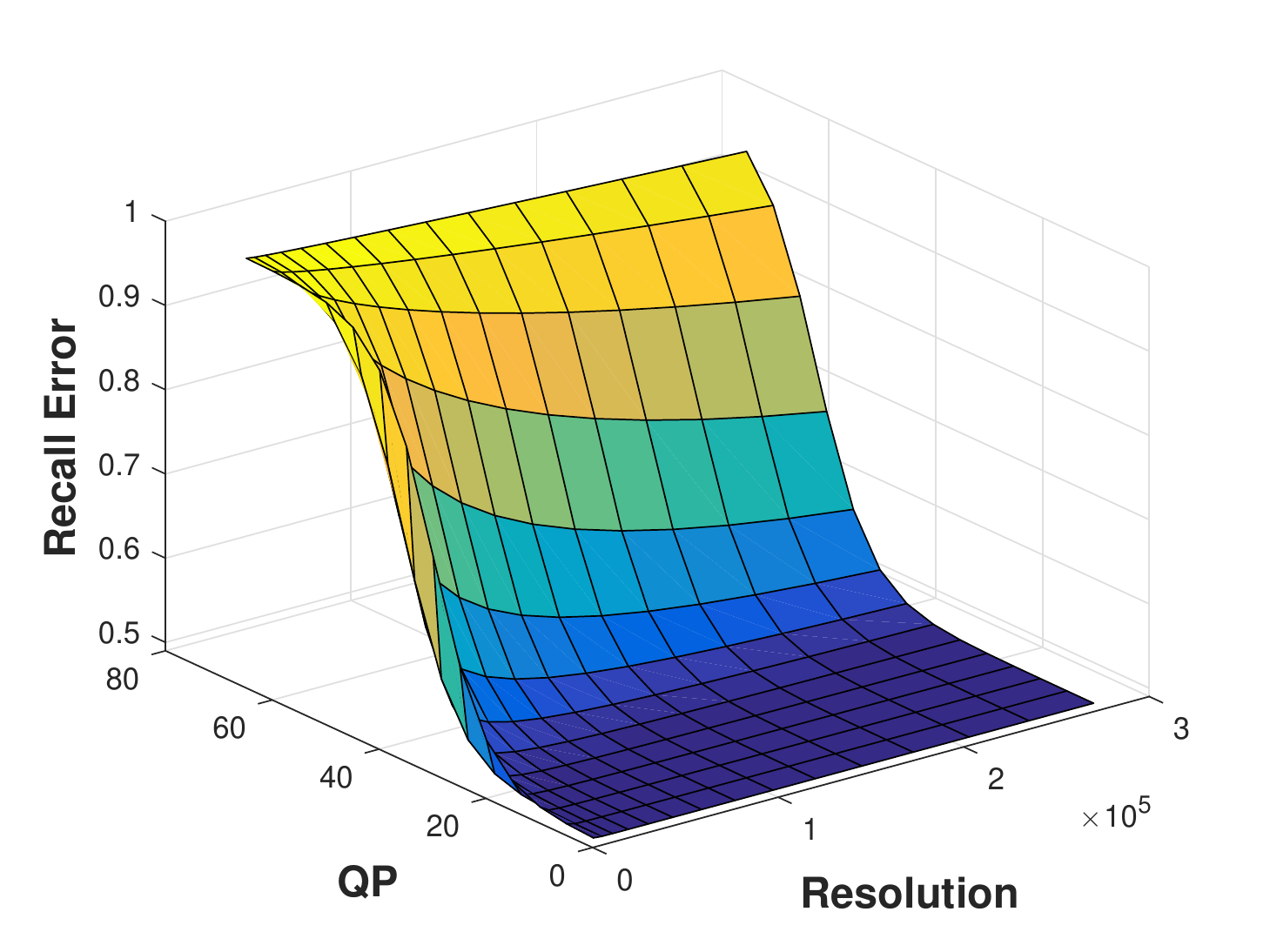}
		\endminipage\hfill
	}
	\subfigure[PCA (DISFA)]
	{
		\minipage{0.23\textwidth}
		\label{fig:DISFAqp3dqp_experimental}
		\includegraphics[width=\linewidth, trim={15 0 30 0}, clip]{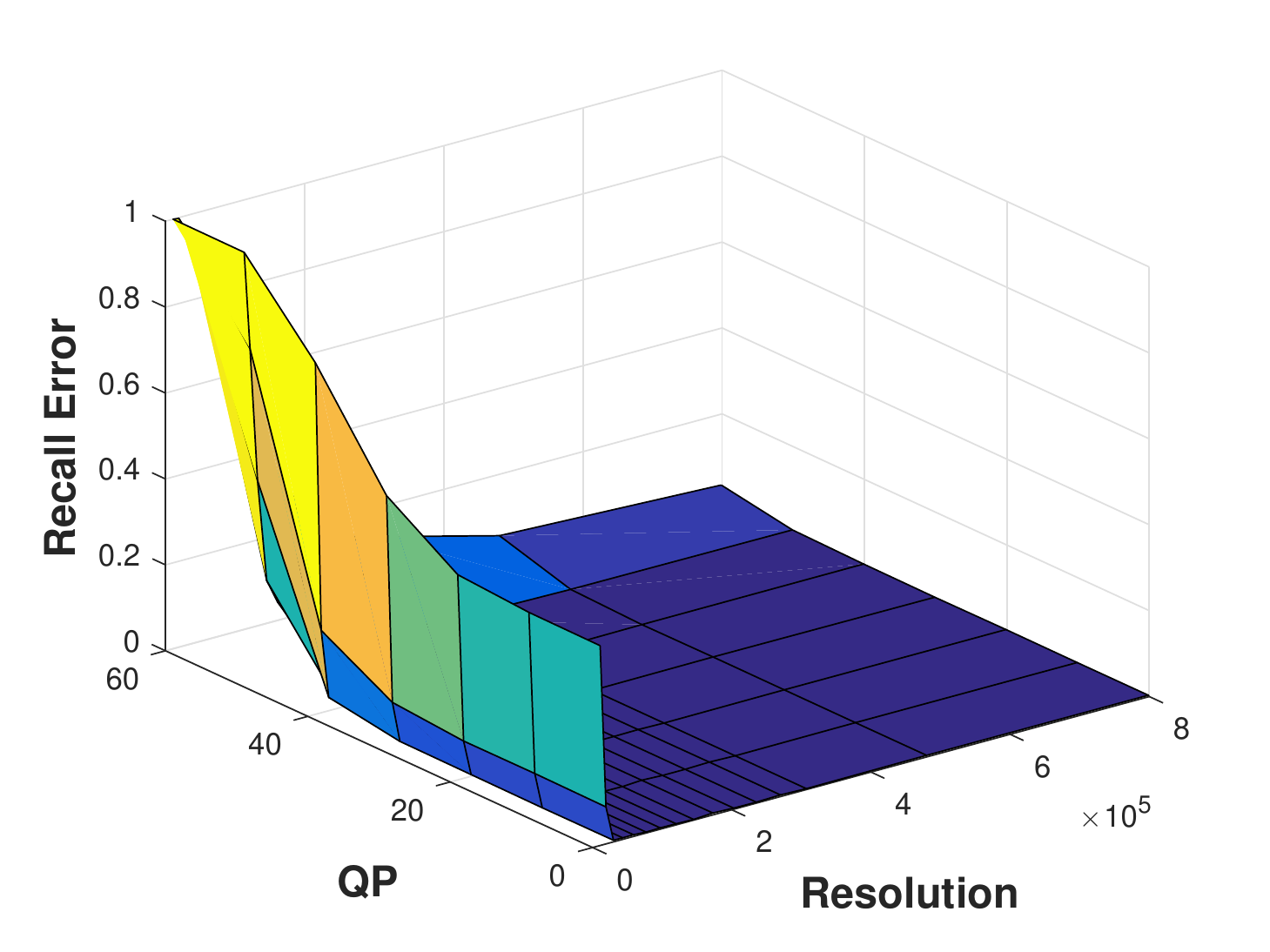}
		\endminipage\hfill
	}
	\subfigure[QRMODA (DISFA)]
	{
		\minipage{0.23\textwidth}
		\label{fig:DISFAqp3dqp_model}
		\includegraphics[width=\linewidth, trim={15 0 30 0}, 	clip]{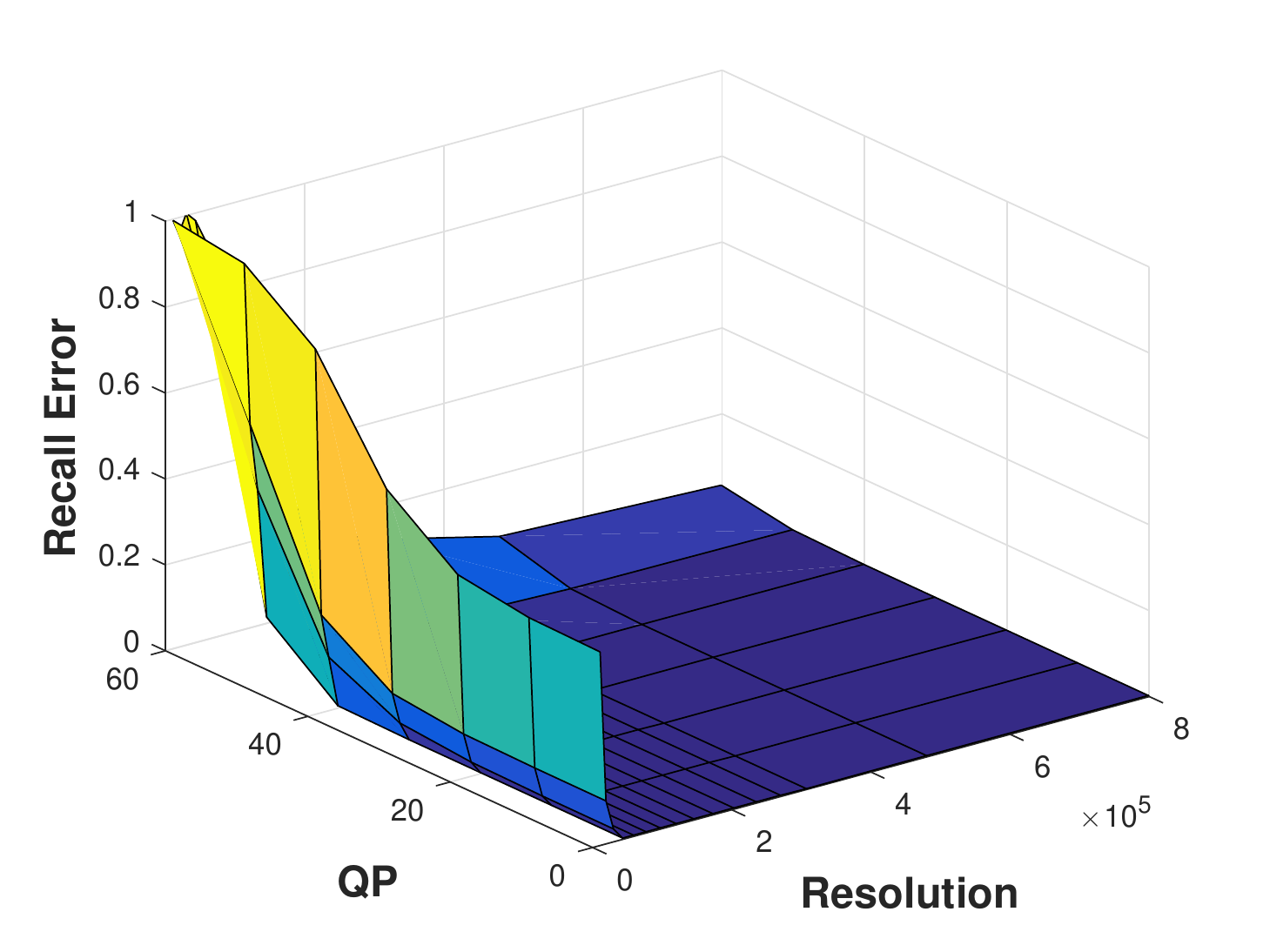}
		\endminipage\hfill
	}
	\caption{Validation and Analysis of QRMODA}
	\label{fig:DISFAqp}
\end{figure*}

\begin{figure*}[!ht]
	\centering
	\subfigure[$600\times 450$ Honda/UCSD]
	{
		\minipage{0.23\textwidth}
		\includegraphics[width=\linewidth, trim={15 0 30 0}, clip]{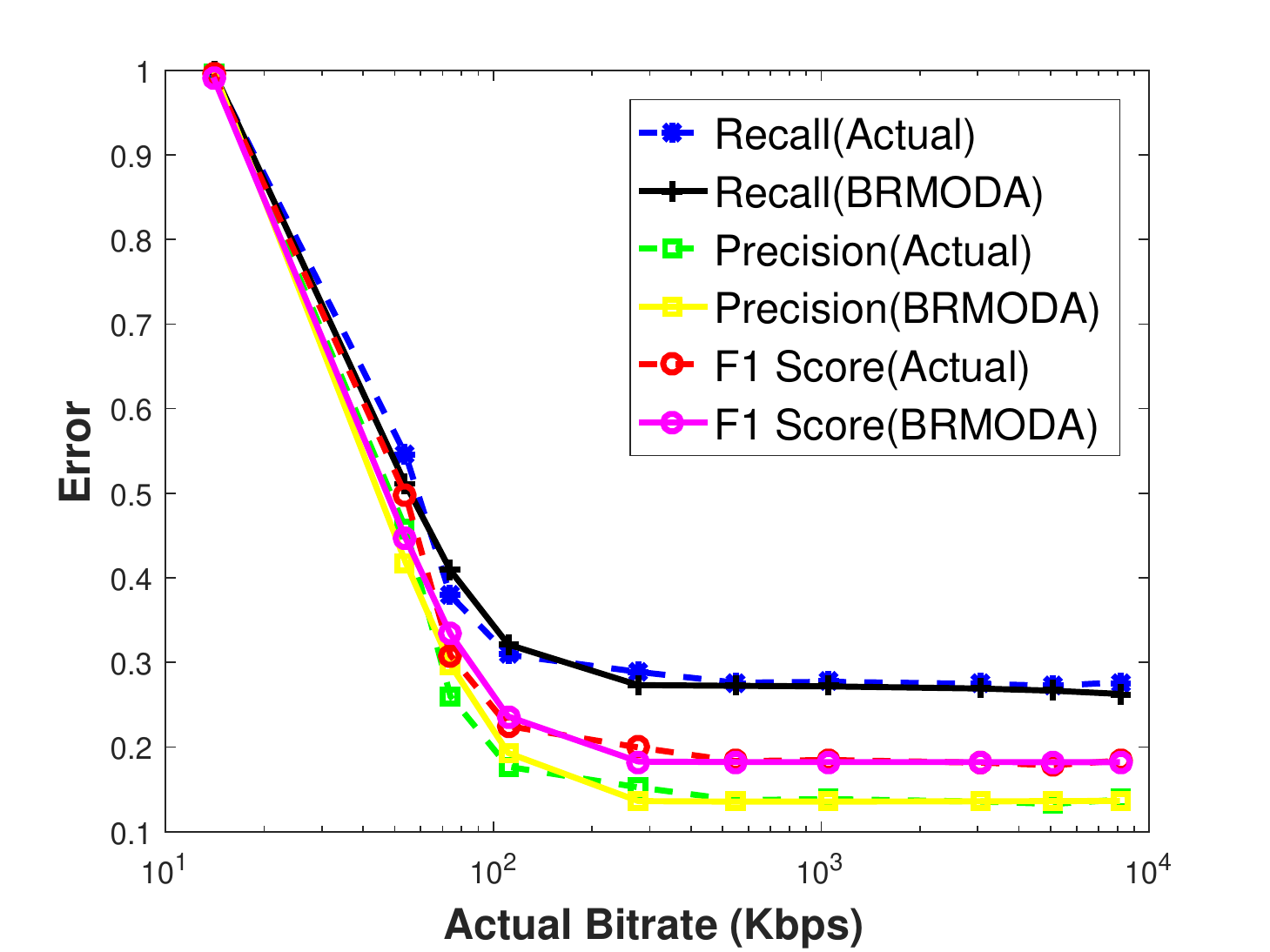}
		\label{fig:f1score}
		\endminipage\hfill
	}
	\subfigure[$600\times 450$  ($R^2$: 0.997, 0.994) Honda/UCSD]
	{
		\minipage{0.23\textwidth}
		\label{fig:neuralbr600}
		\includegraphics[width=\linewidth, trim={15 0 30 0}, clip]{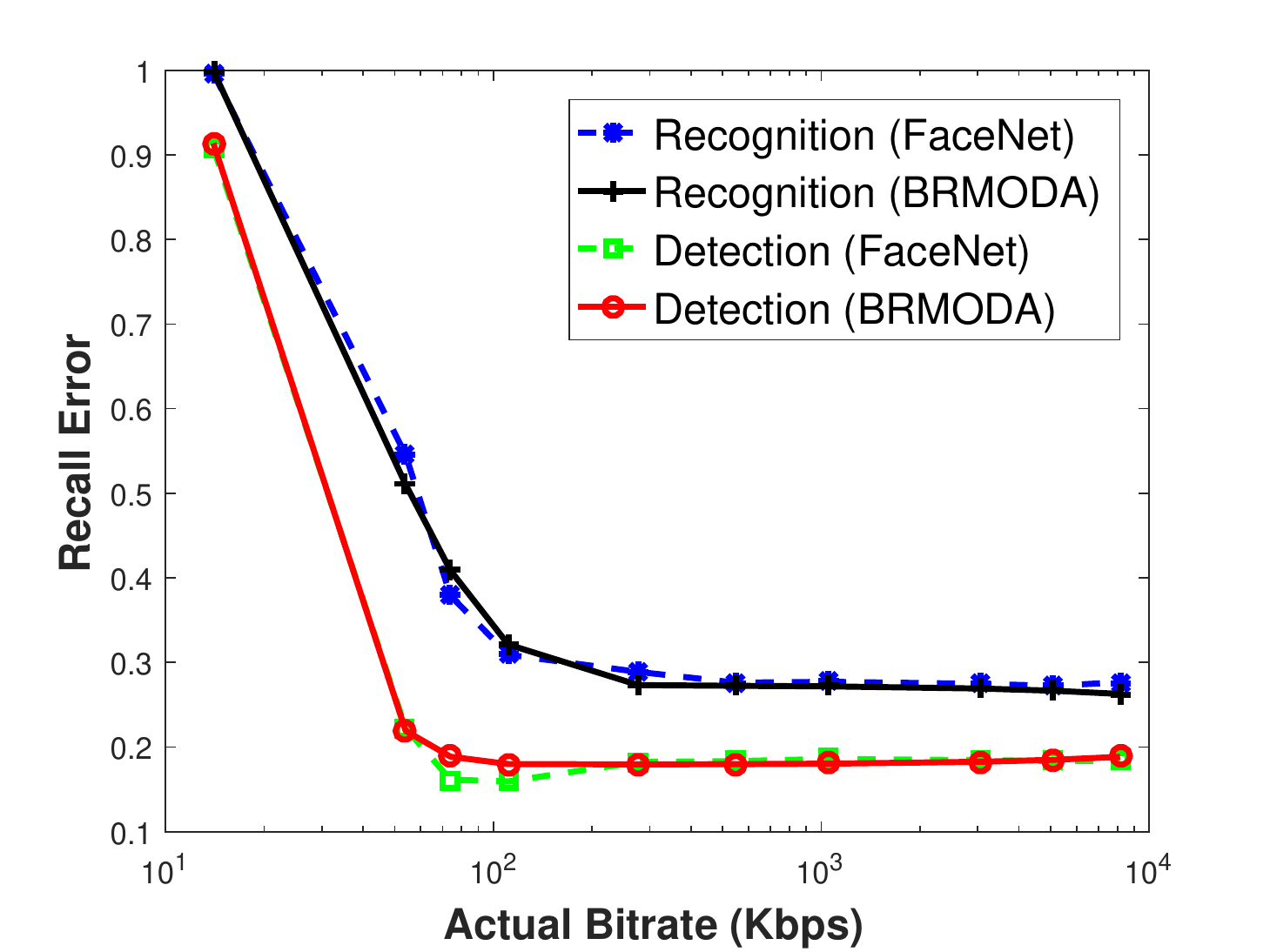}
		\endminipage\hfill
	}
	\subfigure[$480\times 360$  ($R^2$: 0.998, 0.994) Honda/UCSD]
	{
		\minipage{0.23\textwidth}
		\label{fig:neuralbr480}
		\includegraphics[width=\linewidth, trim={15 0 30 0}, clip]{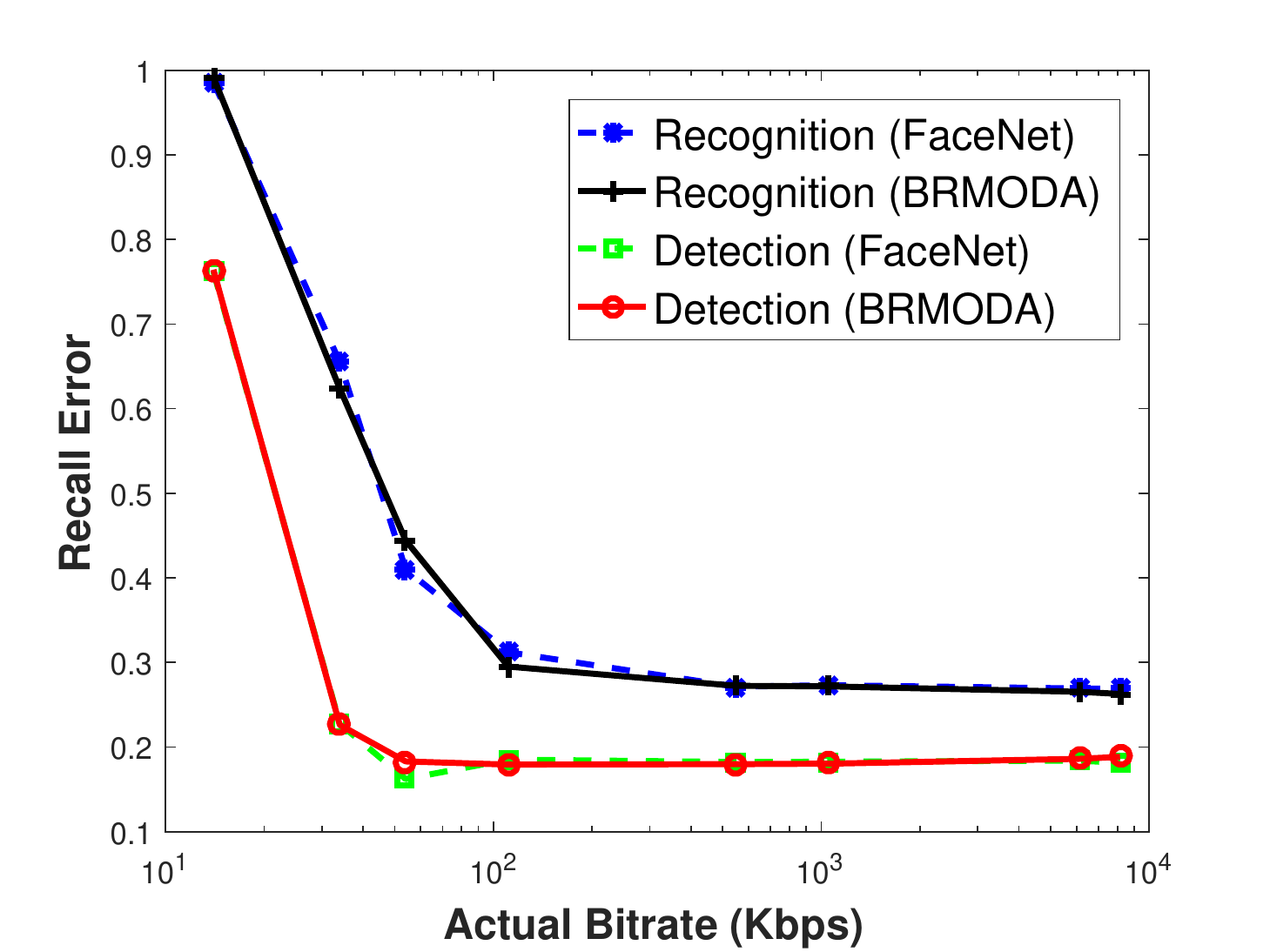}
		\endminipage\hfill
	}
	\subfigure[$360\times 270$  ($R^2$: 0.994, 0.998) Honda/UCSD]
	{
		\minipage{0.23\textwidth}
		\label{fig:neuralbr360}
		\includegraphics[width=\linewidth, trim={15 0 30 0}, clip]{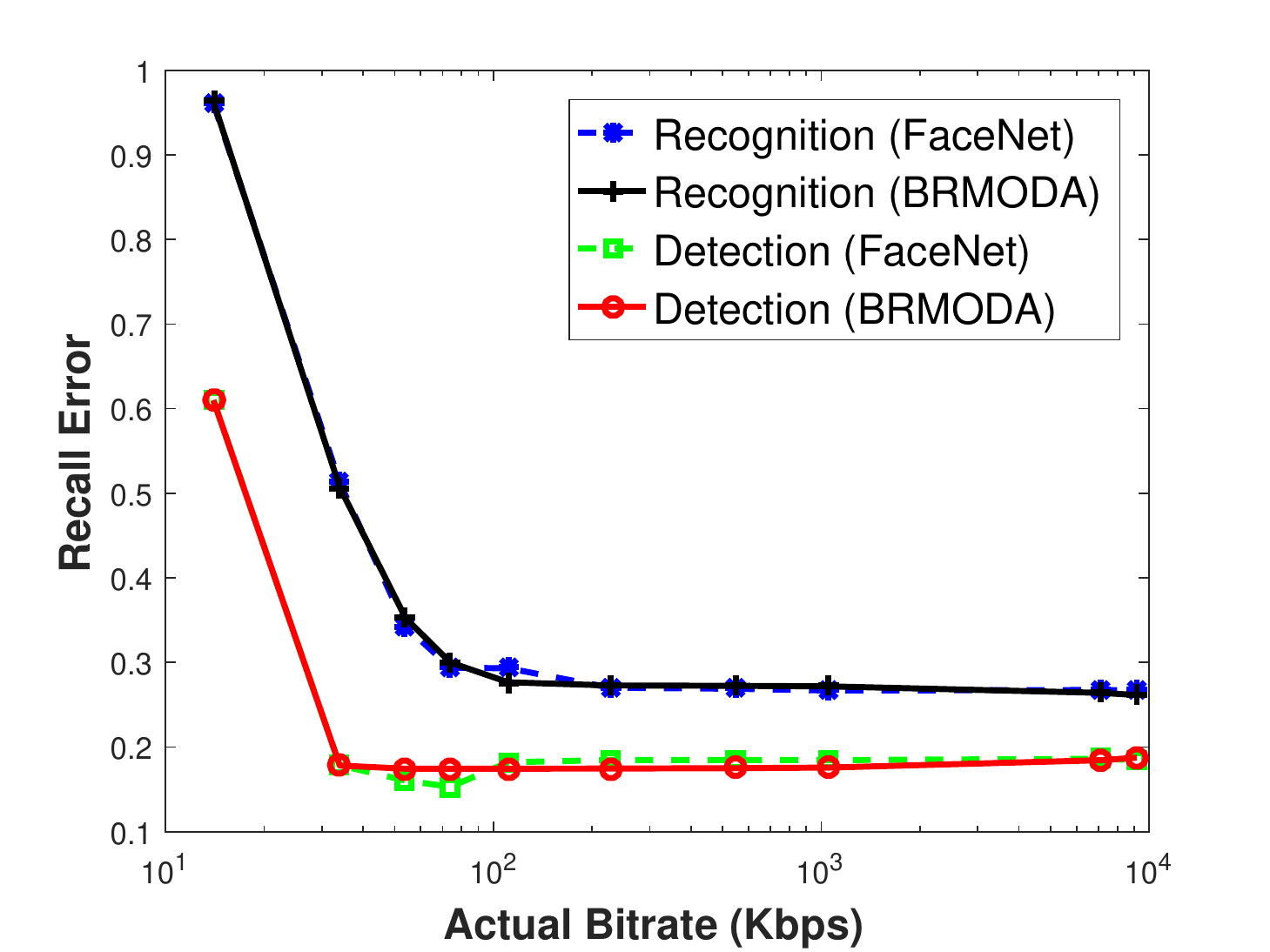}
		\endminipage\hfill
	}
	\subfigure[$240\times 180$  ($R^2$: 0.992, 0.997) Honda/UCSD]
	{
		\minipage{0.23\textwidth}
		\label{fig:neuralbr240}
		\includegraphics[width=\linewidth, trim={15 0 30 0}, clip]{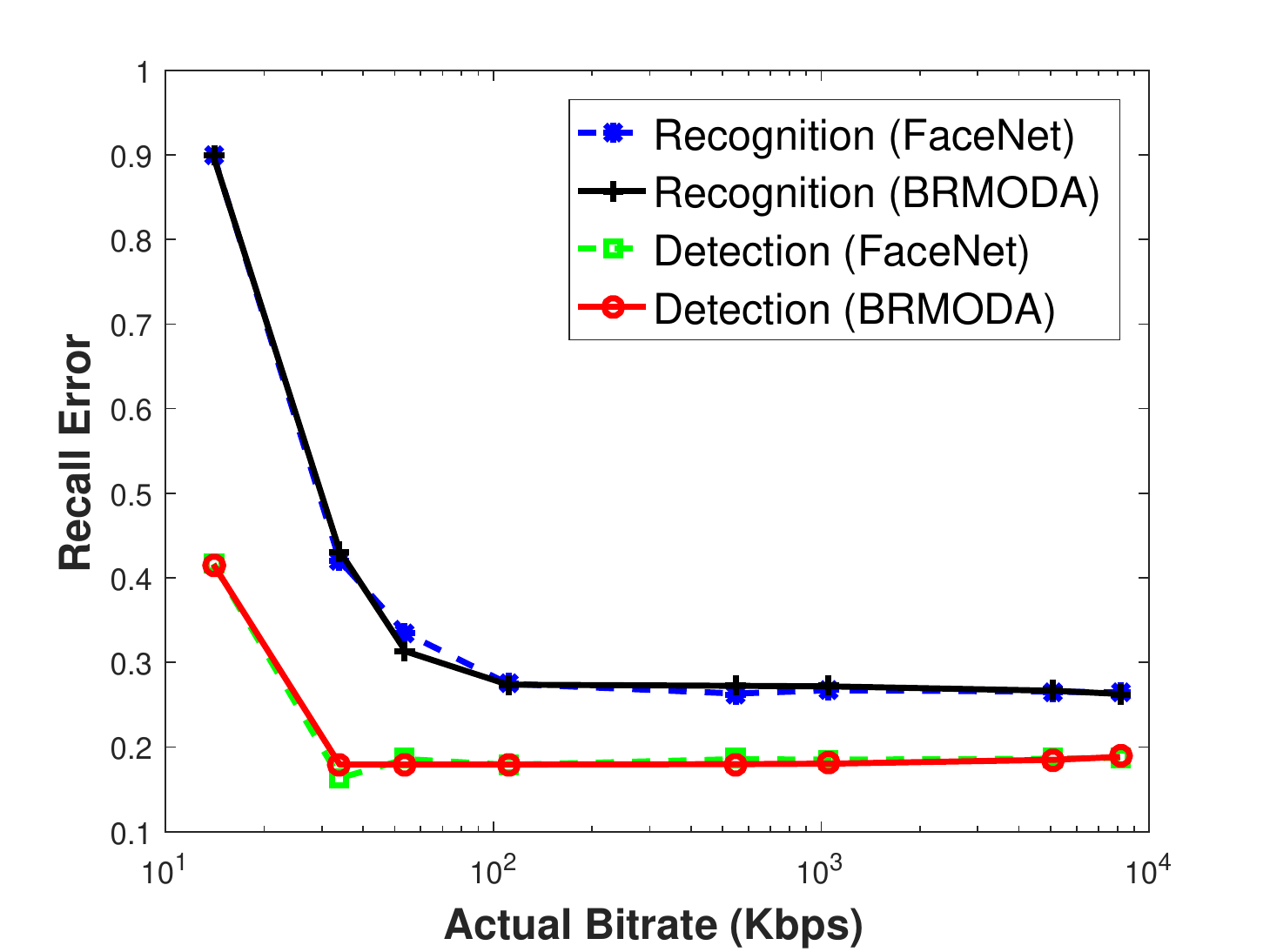}
		\endminipage\hfill
	}
	\subfigure[$400\times 300$  ($R^2$: 0.894, 0.999) DISFA]
	{
		\minipage{0.23\textwidth}
		\label{fig:neuralbrdisfa400}
		\includegraphics[width=\linewidth, trim={15 0 30 0}, clip]{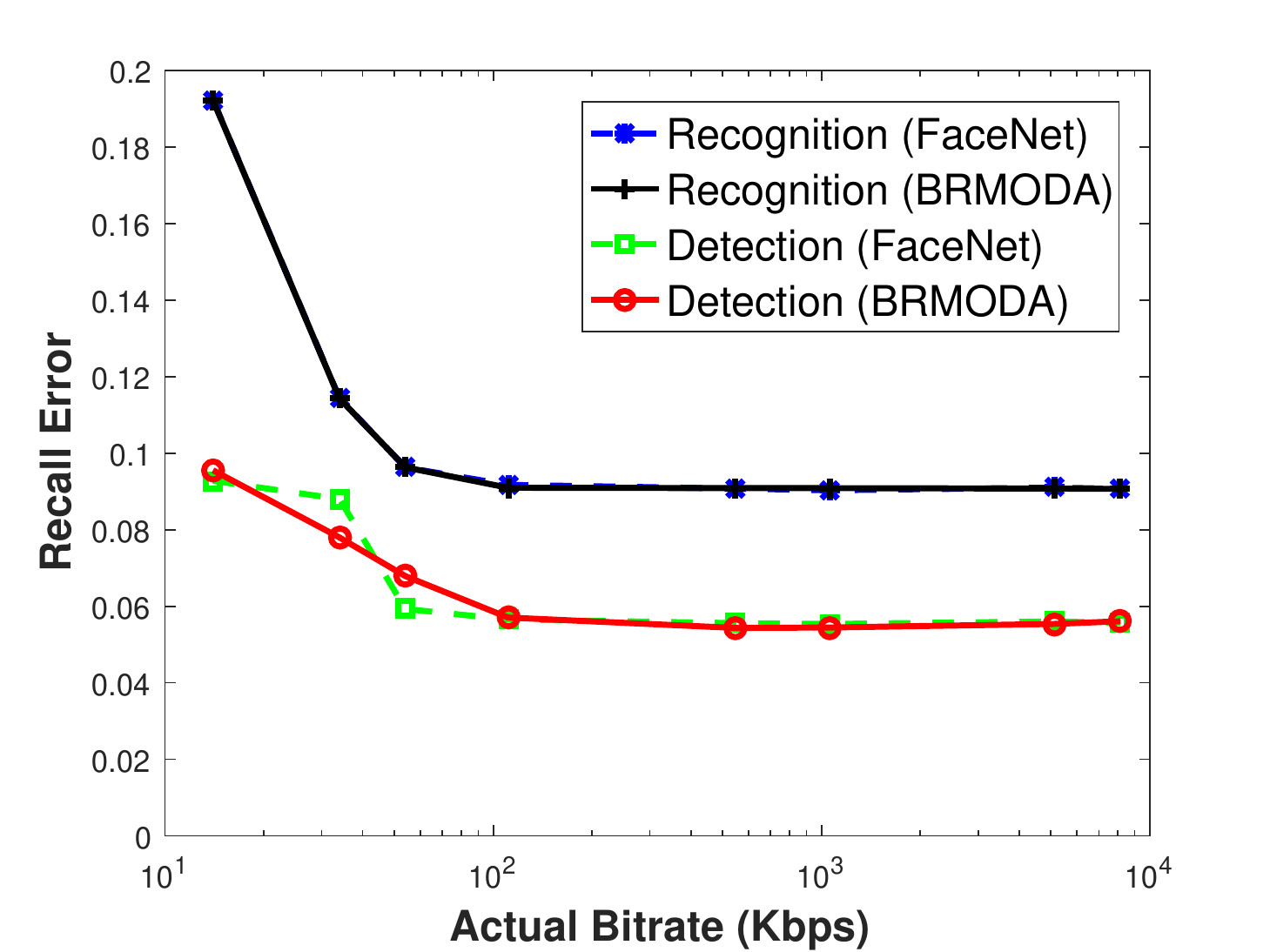}
		\endminipage\hfill
	}
	\subfigure[$240\times 180$ Resolution ($R^2$: 0.995, 0.992) DISFA]
	{
		\minipage{0.23\textwidth}
		\label{fig:neuralbrdisfa240}
		\includegraphics[width=\linewidth, trim={15 0 30 0}, clip]{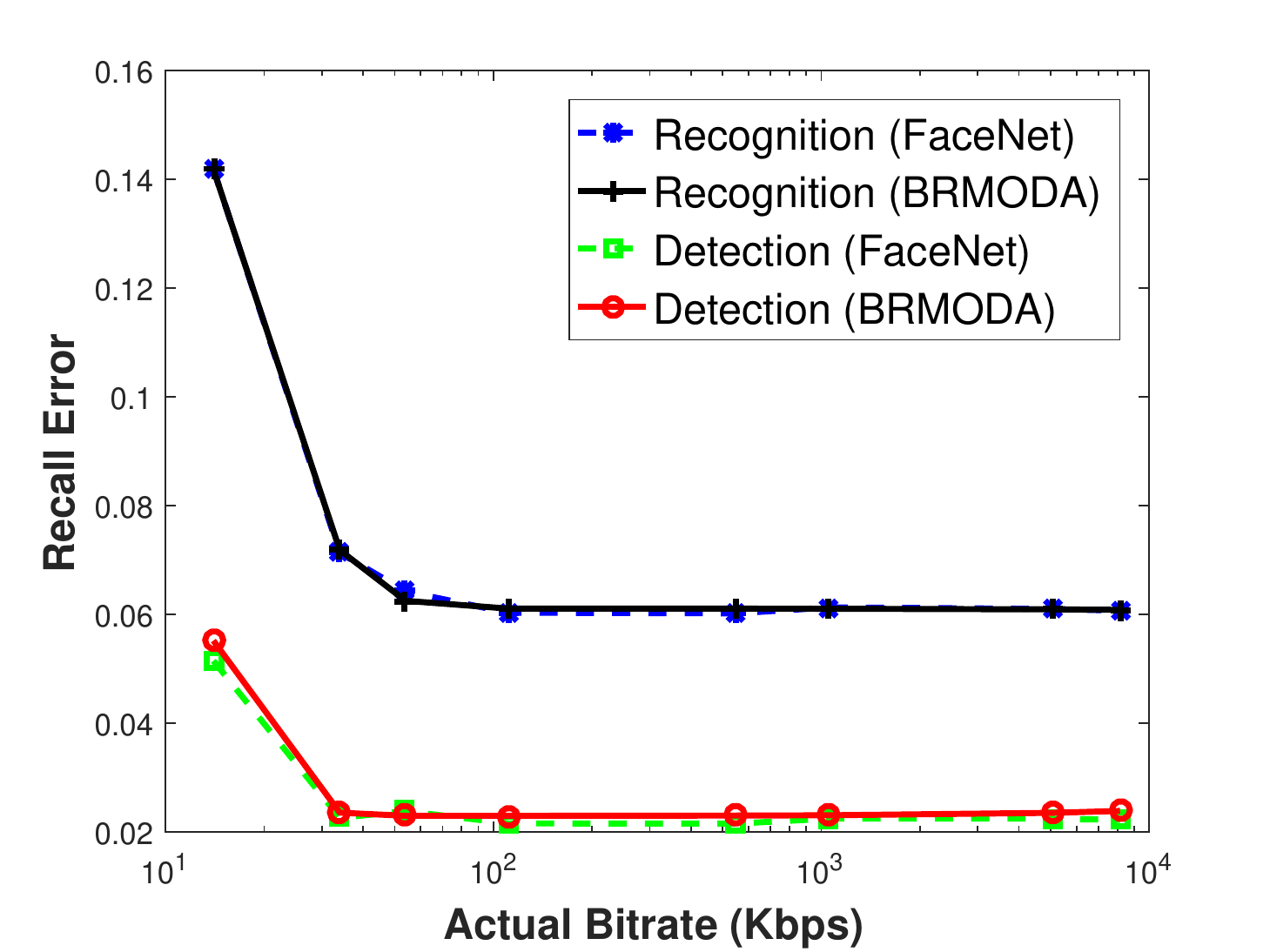}
		\endminipage\hfill
	}
	\subfigure[$120\times 90$ ($R^2$: 0.991, 0.997) DISFA]
	{
		\minipage{0.23\textwidth}
		\label{fig:neuralbrdisfa120}
		\includegraphics[width=\linewidth, trim={15 0 30 0}, clip]{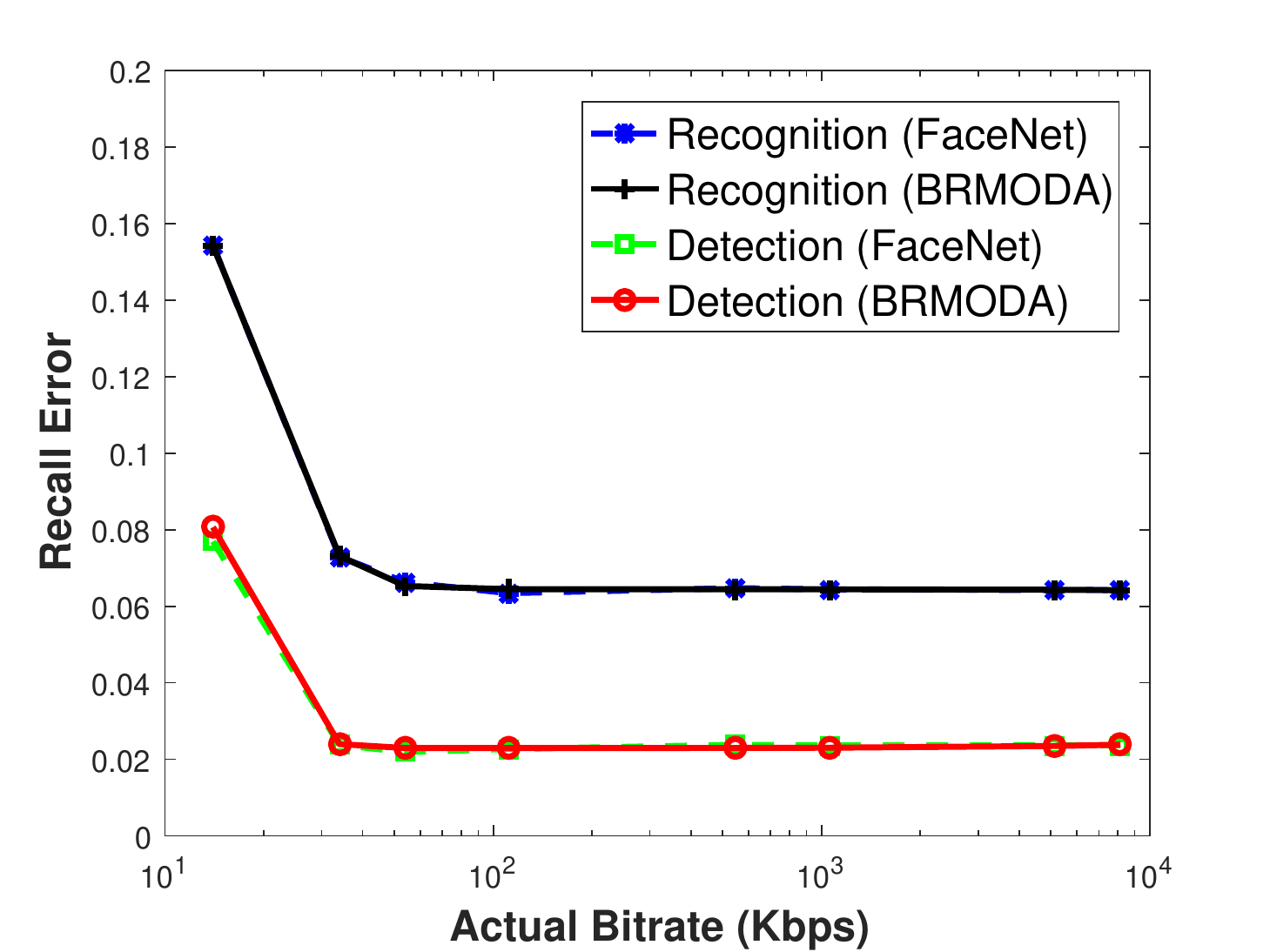}
		\endminipage\hfill
	}
	\centering	
	\subfigure[$600\times 450$ ($R^2$: 0.998, 0.994) Honda/UCSD]
	{
		\minipage{0.23\textwidth}
		\label{fig:errorbrboth600}
		\includegraphics[width=\linewidth, trim={15 0 30 0}, clip]{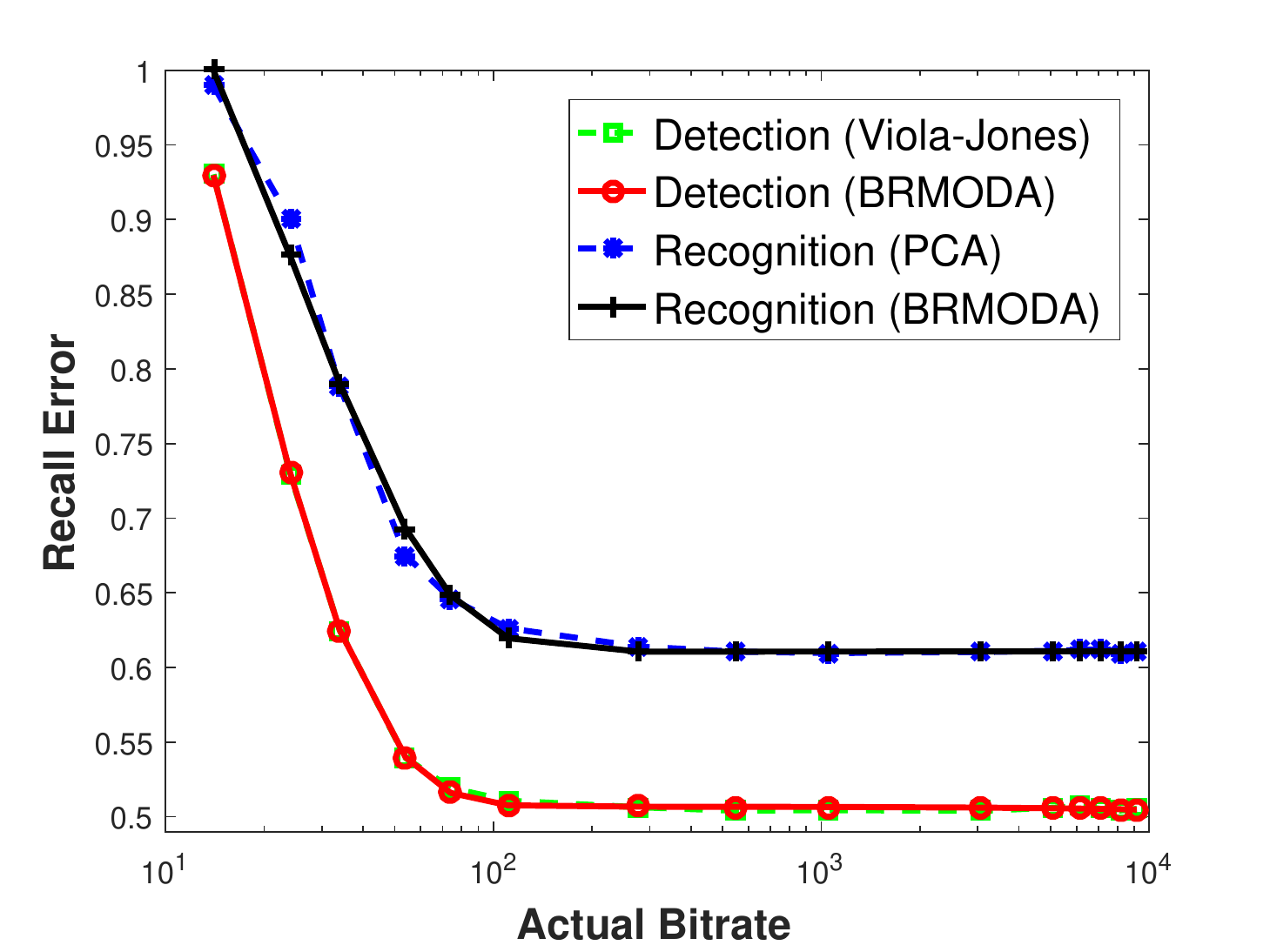}
		\endminipage\hfill
	}
	\subfigure[$520\times 390$ Resolution ($R^2$: 0.997, 0.981) Honda/UCSD]
	{
		\minipage{0.23\textwidth}
		\label{fig:errorbrboth520}
		\includegraphics[width=\linewidth, trim={15 0 30 0}, clip]{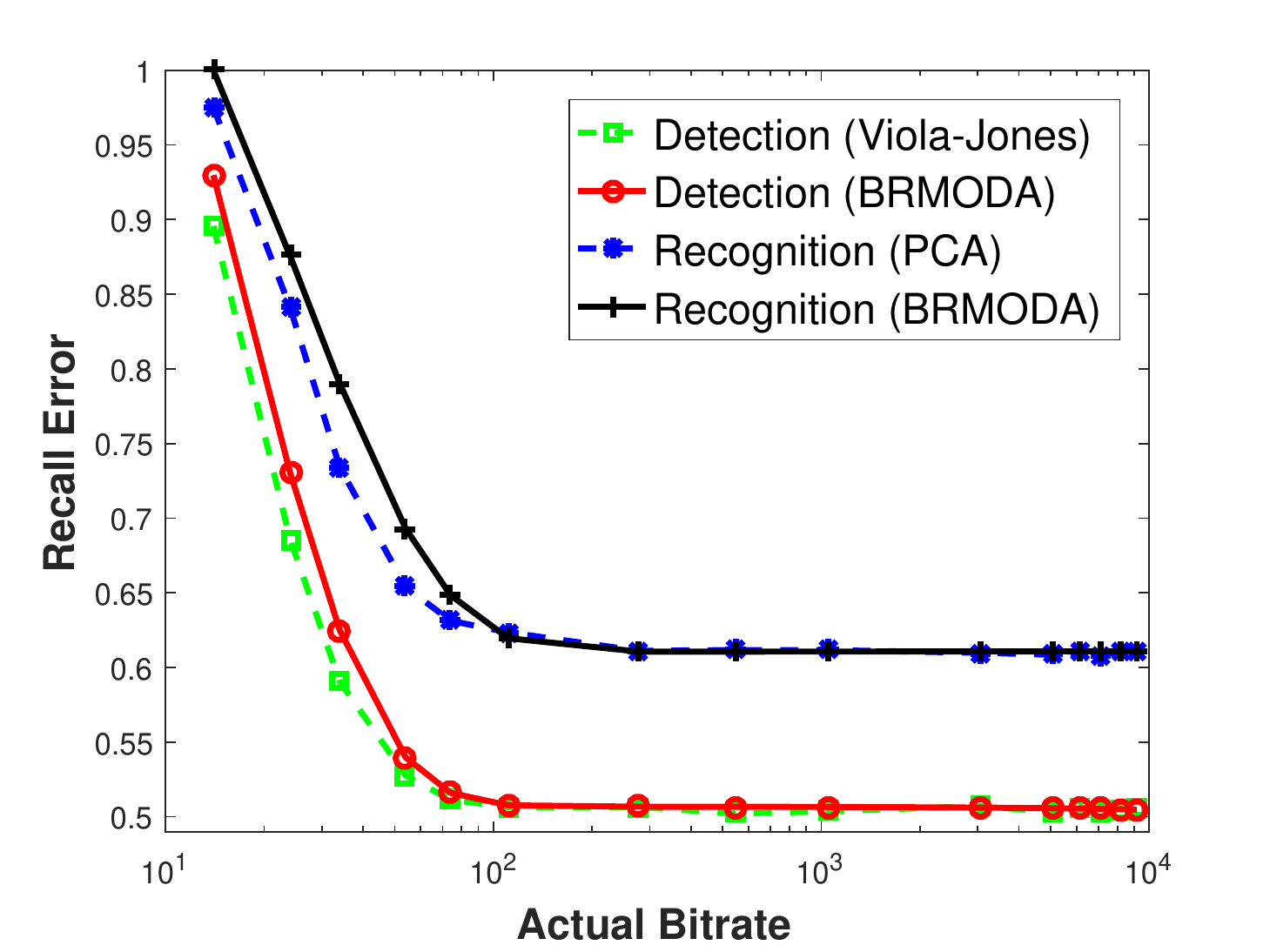}
		\endminipage\hfill
	}
	\subfigure[$400\times 300$ Resolution ($R^2$: 0.991, 0.971) Honda/UCSD]
	{
		\minipage{0.23\textwidth}
		\label{fig:errorbrboth400}
		\includegraphics[width=\linewidth, trim={15 0 30 0}, clip]{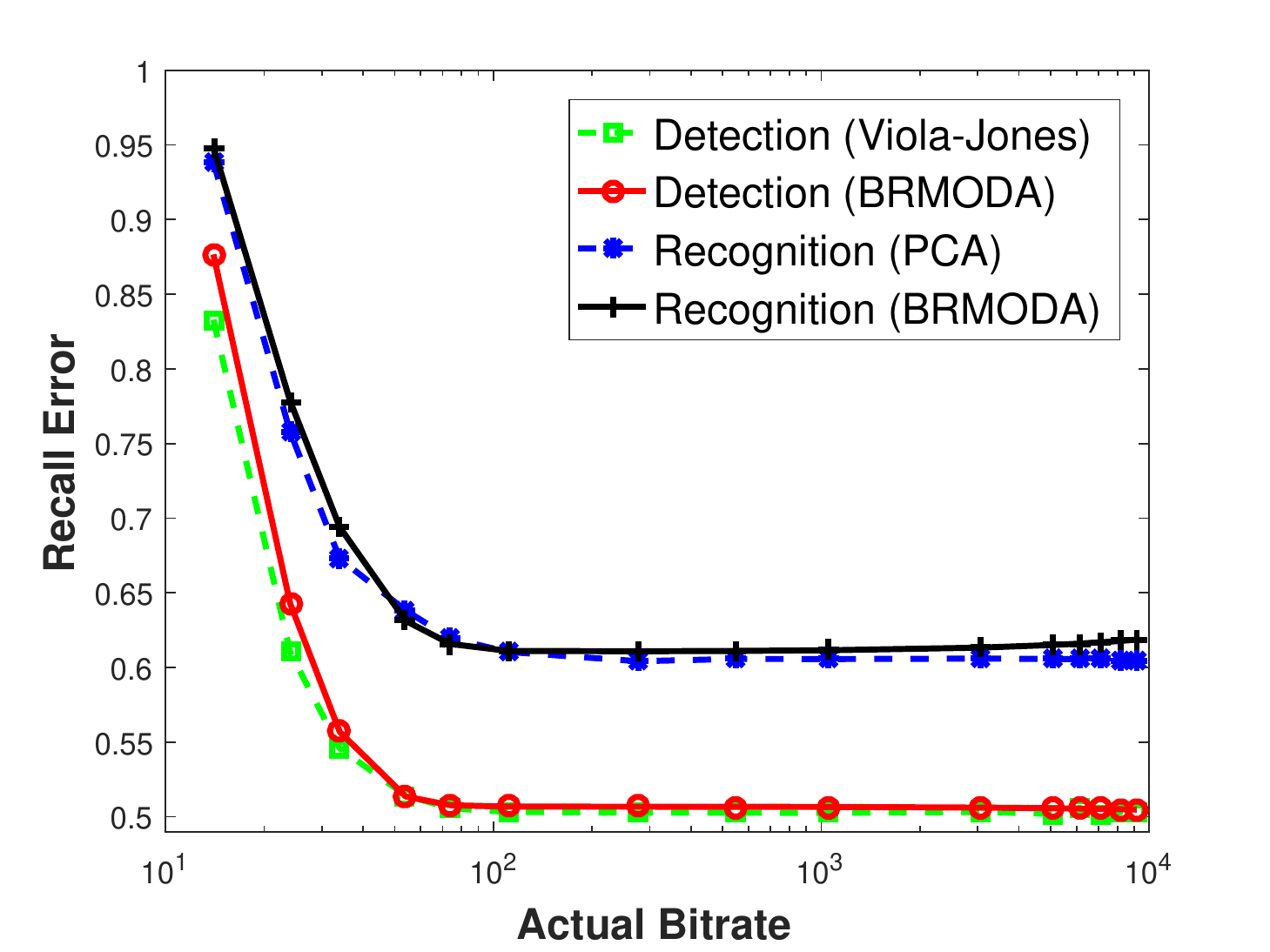}
		\endminipage\hfill
	}
	\subfigure[$200\times 150$ ($R^2$: 0.982, 0.936) Honda/UCSD]
	{
		\minipage{0.23\textwidth}
		\label{fig:errorbrboth200}
		\includegraphics[width=\linewidth, trim={15 0 30 0}, clip]{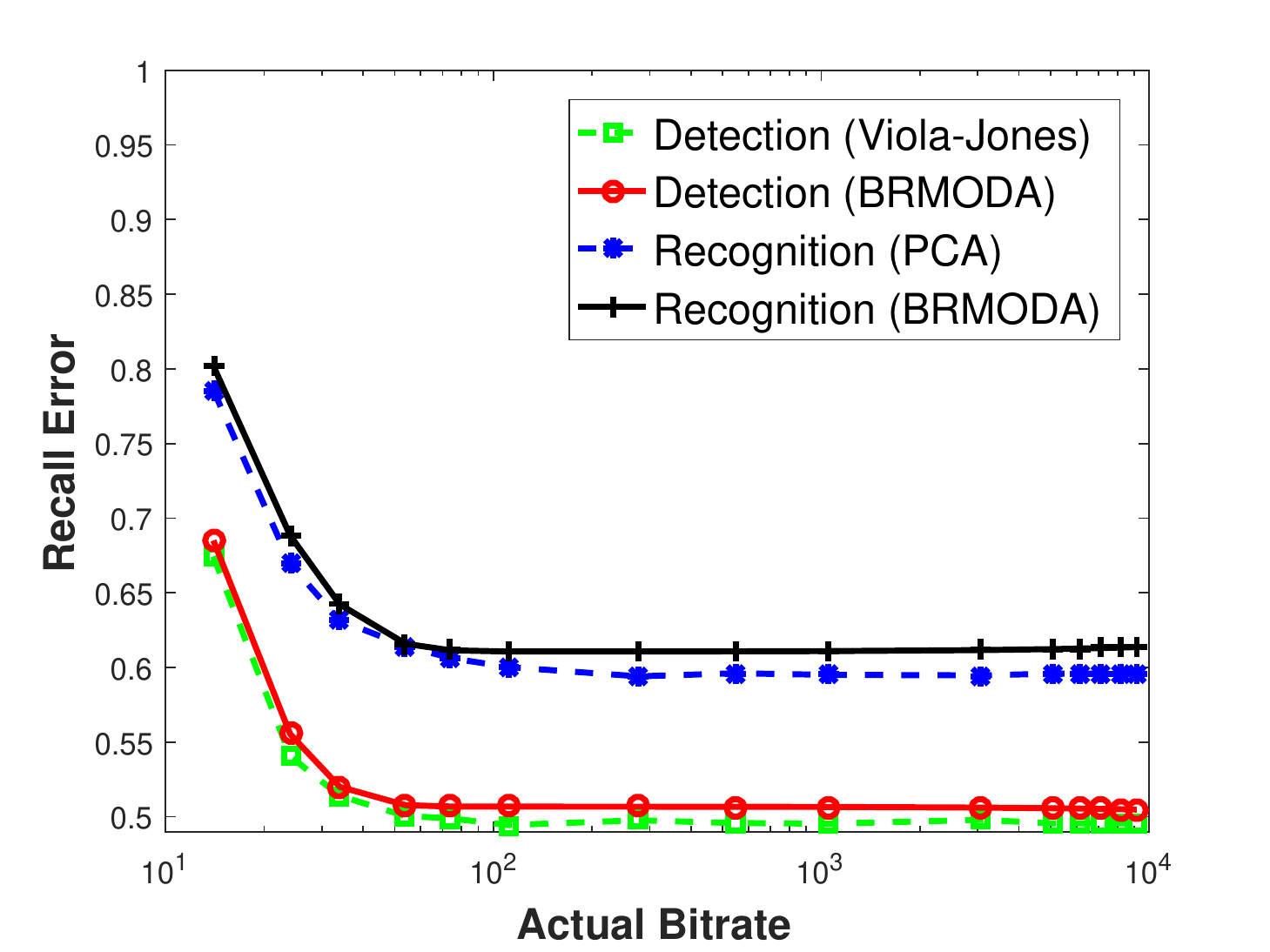}
		\endminipage\hfill
	}
	\subfigure[$560\times 420$ ($R^2$: 0.999, 0.999) DISFA]
	{
		\minipage{0.23\textwidth}
		\label{fig:DISFAerrorbr560}
		\includegraphics[width=\linewidth, trim={15 0 30 0}, clip]{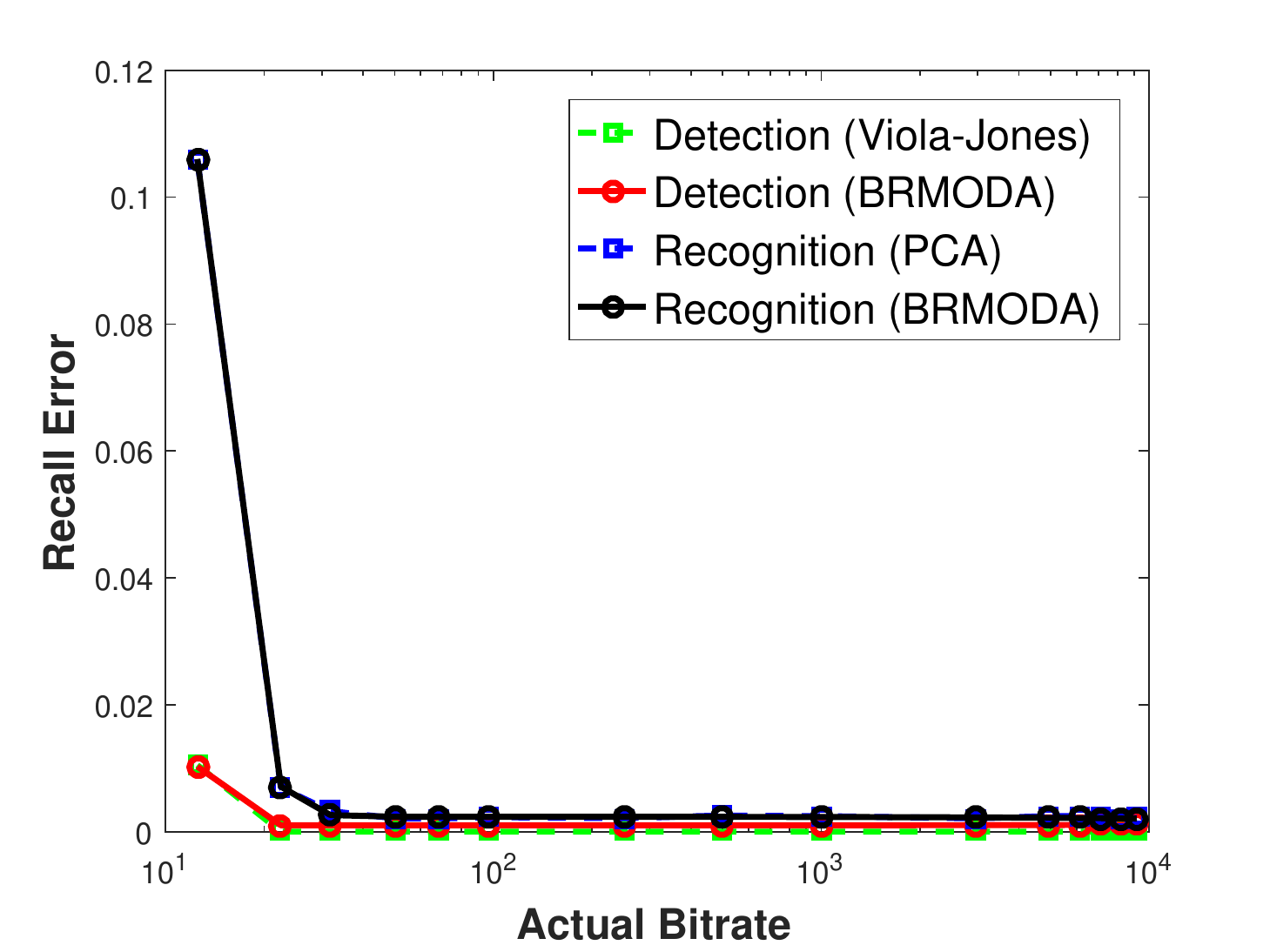}
		\endminipage\hfill
	}
	\subfigure[$480\times 360$ Resolution ($R^2$: 0.9091, 0.999) DISFA] 
	{
		\minipage{0.23\textwidth}
		\label{fig:DISFAerrorbr480}
		\includegraphics[width=\linewidth, trim={15 0 30 0}, clip]{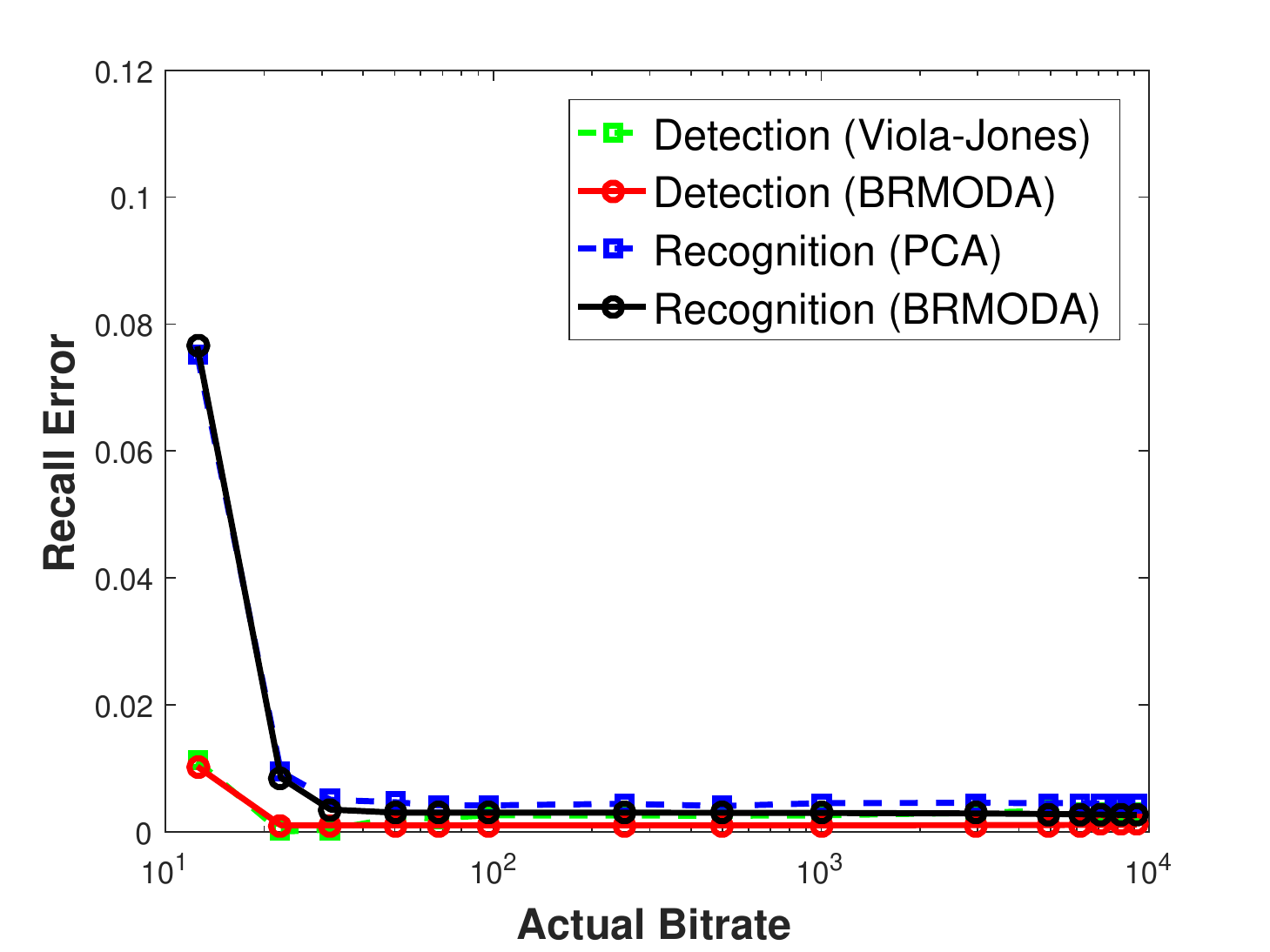}
		\endminipage\hfill
	}
	\subfigure[$360\times 270$ Resolution ($R^2$: 0.9846, 0.997) DISFA] 
	{
		\minipage{0.23\textwidth}
		\label{fig:DISFAerrorbr360}
		\includegraphics[width=\linewidth, trim={15 0 30 0}, clip]{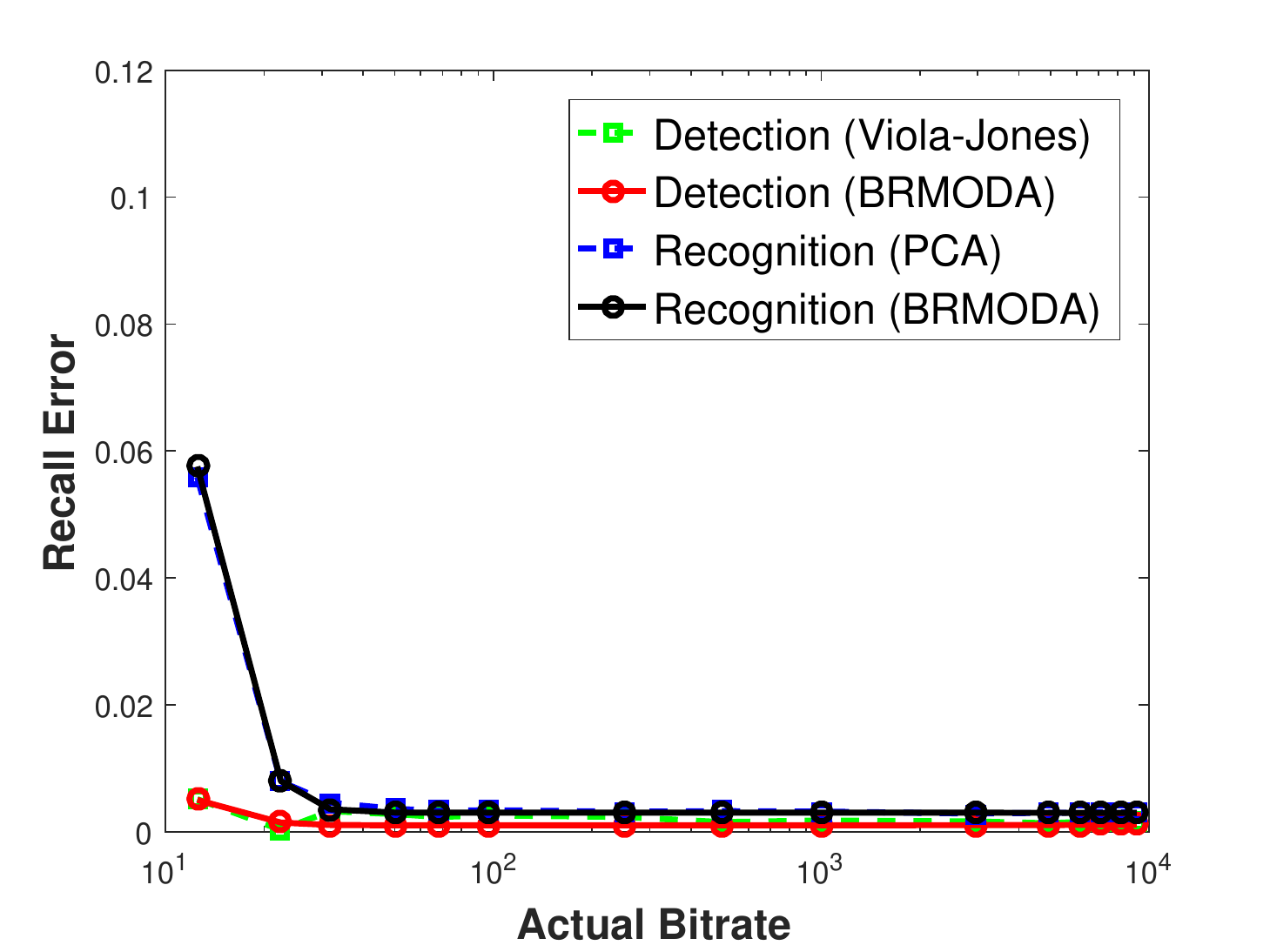}
		\endminipage\hfill
	}
	\subfigure[$280\times 210$ ($R^2$: 0.9458, 0.996) DISFA] 
	{
		\minipage{0.23\textwidth}
		\label{fig:DISFAerrorbr280}
		\includegraphics[width=\linewidth, trim={15 0 30 0}, clip]{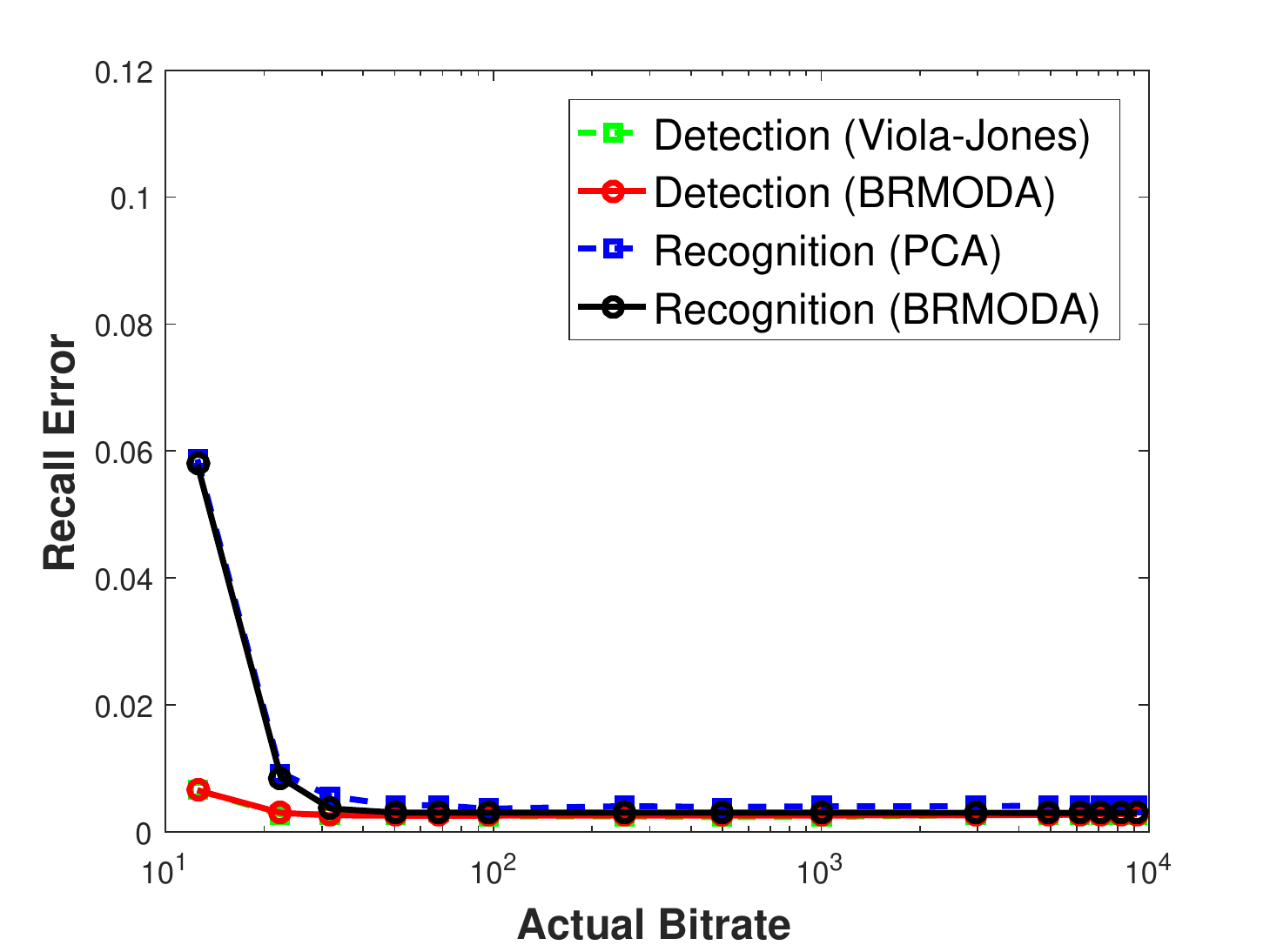}
		\endminipage\hfill
	}
	\caption{Validation and Analysis of BRMODA [\ref{fig:f1score}-\ref{fig:neuralbrdisfa120}: Neural-Based, \ref{fig:errorbrboth600}-\ref{fig:DISFAerrorbr280}: Statistical-Based]
	}
	\label{fig:DISFAbr}
\end{figure*}

\subsection{Result Presentation and Analysis}

For each model, we show the results for experiments performed using two methods of face detection/recognition (neural and statistical based) and utilizing three different datasets for QRMODA and two video datasets for BRMODA (since bitrate adaptation is not applicable to the image dataset). For each set of experiments, we present two subfigures that demonstrate the extremes of the resolution adaptations considered.

We validate QRMODA in terms of detection and recognition sensitivity at different spatial resolutions, as shown in Figure \ref{fig:DISFAqp}.
Subfigures \ref{fig:neuralqp600} through \ref{fig:neuralqp280} show the recall error with respect to $Q_p$ variations for the Honda/UCSD dataset. FaceNet is used for both face detection and recognition tasks.
Similarly, Subfigures \ref{fig:error600} to \ref{fig:error240} show QRMODA's robustness to a change in the detection/recognition methods, when Viola-Jones algorithm is used for face detection, and PCA is used for recognition. 
Subfigures \ref{fig:DISFAqp480} and \ref{fig:DISFAqp280} demonstrate the validation using a different dataset (DISFA).
We also validate QRMODA using state-of-the-art (according to a recent evaluation \cite{soalfw}) 
results on LFW as shown in Subfigures \ref{fig:lfw50} and \ref{fig:lfw200}. 
Additionally, a 3D graph of PCA vs. QRMODA is shown in 
Subfigures \ref{fig:error3dqp_experimental} - \ref{fig:DISFAqp3dqp_model} using Honda/UCSD and DISFA, respectively. 
The latter subfigures demonstrate the entire model behavior with respect to simultaneous variations in both encoding parameters (i.e. $Q_p$ and the Resolution) represented by $x$ and $y$ axes, respectively. The recall error is color-coded over the $z$-axis. 
The results demonstrate that our proposed model is highly accurate.

The recall error increases slowly with  $Q_p$ up to a critical point,
represented by the   Sigmoid's midpoint of the Logistic function.
After that point, the error increases sharply with $Q_p$  until it becomes $100\%$.
The lower bound for face detection error is about $0.2$ for the Honda/UCSD and approximately $0$ for
the DISFA. Additionally,  both
the detection and recognition recall rates in DISFA are much higher than
those in Honda/UCSD. 
This difference in threshold levels is because of the variation in video contents,
which contain fewer frontal face poses in the Honda/UCSD
than those in  DISFA. The pose angle is recognized as an important factor in detection and recognition.
Table \ref{table:QRMODACNN}
lists some of the constants used in this study.

Figure \ref{fig:DISFAbr} validates BRMODA in terms of detection and recognition recall errors at different resolutions, using neural-based and the statistical-based methods, respectively. 
Each subfigure shows the normalized recall error versus the actual bitrate for selected resolutions.
As the target bitrates may not be achieved precisely by the encoder, we report the actual bitrates, 
which are depicted in the figures on a logarithmic scale because
of the wide range of considered bitrates in our experiments.
The results demonstrate that the model is highly accurate in terms of the calculated $R^2$.
Remarkably, both models can be applied to other accuracy measures as well, such as precision and F1-score.
Subfigure \ref{fig:f1score} demonstrates how BRMODA can be applied to all the different metrics.
The recall is inversely proportional to
the actual bitrate achieved due to the negative value of $c'_3$. 
This behavior varies with spatial resolution variation because high
resolution videos require high bitrates, thereby produce high
errors when low bitrates are  imposed. 
For this reason, the recall error value starts at larger values in Subfigures
\ref{fig:neuralbr600}, \ref{fig:neuralbr480}, \ref{fig:neuralbrdisfa400}, \ref{fig:errorbrboth600}, \ref{fig:errorbrboth520}, and \ref{fig:DISFAerrorbr560} than those in
Subfigures \ref{fig:neuralbr360}, \ref{fig:neuralbr240}, \ref{fig:neuralbrdisfa240}, \ref{fig:neuralbrdisfa120}, \ref{fig:errorbrboth200}, \ref{fig:DISFAerrorbr360}, and \ref{fig:DISFAerrorbr280}, respectively.  
\section{Discussion}

\label{sec:discussion}

The actual recall rate depends on different factors, including
the subject's pose angle and inter-ocular distance in pixels.
Both these factors depend on the camera placement and settings (zoom, pan, and tilt). 
Other factors are related to the environment, such as lighting and potential occlusion
by other subjects or objects in the scene. 
A further factor is the face recognition algorithm being used. As shown in Section \ref{sec:validation}, 
CNN-based face recognition achieves the highest recall. This is not only due to the merits of deep-learning but also the ability of FaceNet to detect and align frames with side facial poses, whereas the cascade classifiers used in OpenCV fails to do so.

The proposed models characterize the recall error in terms of the main encoding parameters.
The constant values can be determined dynamically upon system calibration.
We recommend that the constants are determined based on  actual videos captured by the
cameras in the (surveillance) site. 
The system can generate different adaptations, and then determine the constants that best fit the model(s). 
For instance, to determine the constants in QRMODA, {\em monotone regression splines} can be employed. 
Standard curve fitting procedures, such as 
{\em reformatting} and {\em redefining} tools can be used. 
Some of these functions are  available off-the-shelf, including the recently released {\em splines2} package implementation in R \cite{splines2}.
\section{Conclusions}
\label{sec:conclusion}

We have proposed two novel models that characterize CV accuracy.
We have conducted extensive experiments involving combinations of video adaptation techniques to assess the effect of video encoding parameters on the system. We have used two greatly distinct video datasets and a large image dataset for validation.
We have validated the models using both CNN and statistical-based methods and reported $R^2$. The results show that both models are valid under all experimental scenarios.
We find it remarkable that the two models apply to greatly
distinct video/image datasets and to both face recognition and detection. The models also apply to both deep learning and statistical-based methods and can be utilized to capture the different CV accuracy metrics (precision, recall, F1-score).
Ultimately, we have discussed the factors impacting the constants of each model.

\bibliographystyle{ieeetr}
\bibliography{main}

\end{document}